\documentclass{article}
\newif\ifarxivpaper
\ifdefined\ArxivVersion
  \arxivpapertrue
\fi

\PassOptionsToPackage{numbers,compress}{natbib}
\usepackage[main,final]{neurips_2026}

\usepackage[utf8]{inputenc}
\usepackage[T1]{fontenc}
\usepackage{float}
\usepackage{hyperref}
\usepackage{url}
\usepackage{booktabs}
\usepackage{multirow}
\usepackage{array}
\usepackage{amsfonts}
\usepackage{amsmath}
\usepackage{amssymb}
\usepackage{microtype}
\usepackage{xcolor}
\usepackage{graphicx}
\usepackage{xspace}
\usepackage[capitalize,noabbrev]{cleveref}

\newif\ifcameraReadyEdits
\cameraReadyEditsfalse
\DeclareRobustCommand{\cameraReadyEdit}[1]{{\ifcameraReadyEdits\color{blue}\fi#1}}
\newcommand{\ours}{CARE\xspace}
\newcommand{\oursLong}{Cross-Question Abstention Reward Estimation\xspace}

\providecommand{\indicator}{\mathbf{1}}
\DeclareMathOperator*{\argmax}{arg\,max}

\newsavebox{\promptboxcontent}

\title{Cross-Question Reliable Reinforcement Learning}

\author{Hector G. Rodriguez \qquad Marcus Rohrbach \\
TU Darmstadt \& hessian.AI, Germany \\[0.6em]
\url{https://github.com/multimodal-ai-lab/CARE}
}

\begin{document}
\raggedbottom

\maketitle
\vspace{-5pt}
\begin{abstract}
\vspace{-7pt}
AI systems operating near the limits of their capabilities can maximize utility while minimizing the risk of error by abstaining or deferring questions they are not confident about to a human or a stronger model.
Recently, large language models (LLMs) have been post-trained to estimate confidence in their own answers.
To achieve this, methods reinforce confidence scores based on the correctness signal to each question-answer pair \emph{individually}, with rewards that encourage assigning confidence 1 to correct answers and 0 to incorrect answers.
However, estimating confidence scores is useful in that it enables selecting only the most confident answers, to maximize the share of solved questions while accommodating the user's risk tolerance.
Assigning confidence 1 to correct answers and 0 to incorrect ones is theoretically optimal, but unattainable in practice and detrimental for selective prediction with a non-zero risk tolerance: confidences pushed to 0 and 1 carry no information about how answers compare to one another.
We therefore introduce \ours (Cross-question Abstention Reward Estimation), which computes rewards over a set of question-answer pairs:
At each of several risk tolerances, the confidences across the set induce an abstention threshold, and each confidence is rewarded for lying on the correct side of it.
Since the threshold itself comes from the set, every verbalized confidence is reinforced relative to other questions' confidence, not against its own binary correctness label.
On math, \ours increases the coverage at 5\% risk (C@5) across MATH-500, GSM8K, and Big-Math-Digits from 0.8\% to 42.4\%, compared to existing pointwise correctness-matching rewards.
When trained on multi-hop Wikipedia QA from HotpotQA, \ours improves C@5 in out-of-distribution general-knowledge, reasoning, and math benchmarks (including CommonsenseQA, GPQA, SimpleQA, and TriviaQA) from 12.3\% to 23.6\%.

\end{abstract}

\begin{figure}[H]
  \centering
  \vspace{-0.5em}
  \begin{minipage}[c]{0.58\linewidth}
    \centering
    \includegraphics[width=\linewidth]{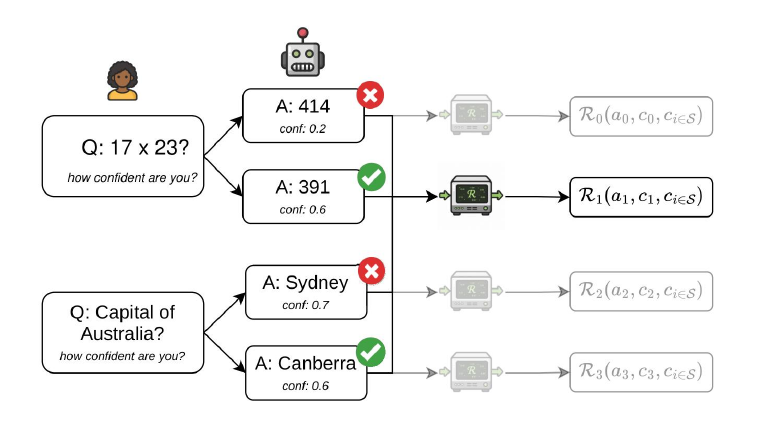}
  \end{minipage}\hfill
  \begin{minipage}[c]{0.40\linewidth}
    \centering
    \includegraphics[width=\linewidth]{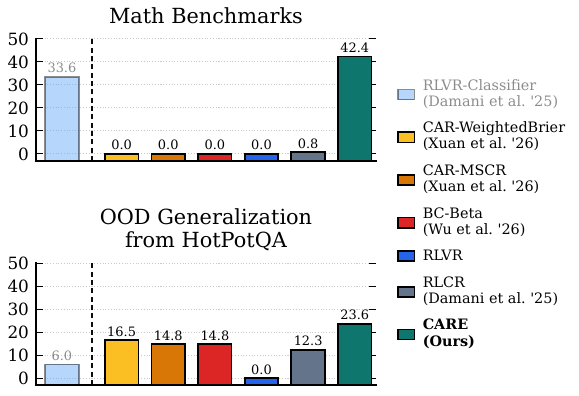}\end{minipage}
  \caption{\textbf{Cross-Question Abstention Reward Estimation (CARE).}
  LLMs are trained to verbalize confidence in their answers.
  Existing pointwise rewards score predicted confidence only against its correctness label, i.e. confidence reward $\mathcal{R}_i(a_i, c_i)$ for answer $a_i$ depends only on that answer's correctness and confidence prediction. \textbf{Left:} \ours uses sets of answers and confidences $\mathcal{S}$ across a batch to reinforce coverage at risk, so the reward $\mathcal{R}_i(a_i, c_i, c_{j\in\mathcal{S}})$ is informed by other confidence predictions $c_{j\in\mathcal{S}}$.
  \textbf{Right:} C@5 on unseen math benchmarks and on HotpotQA-to-OOD transfer.}
  \label{fig:teaser}
  \vspace{-0.5em}
\end{figure}

\section{Introduction}

Large language models are increasingly working in domains where they are far from perfectly accurate, yet can be trusted to solve a substantial fraction of the tasks they encounter. Confidence prediction allows these systems to identify the tasks that are hardest for the model, and defer them to a human expert or a stronger model, therefore increasing automation without sacrificing low risk.
High stakes scenarios where automation is practical yet hardest tasks may be deferred to experts include visual question answering assistance for blind people \citep{bigham2010vizwiz,gurari2018vizwiz,huh2024longform}, autonomous driving vehicles \citep{mcallister2017concrete,ivanovic2022propagating,antonante2023taskaware}, or even coding agents, where orchestrators can decompose a problem, delegate and supervise work from specialized or cheaper worker agents \citep{fourney2024magenticone,anthropic2026subagents,hermes2026delegation,chen2023frugalgpt}.

Autoregressive language models generate answers by sampling from next-token likelihoods, making sequence likelihood a natural source of a model's confidence in its own answer.
However, common state-of-the-art post-training and inference techniques can severely degrade likelihood-based confidence as a predictor of answer correctness \citep{nakkiran2025trained}.
Specifically, \citet{nakkiran2025trained} show that reinforcement learning with verifiable rewards (RLVR) post-training \citep{openai2024o1,deepseek2025r1} moves models away from the likelihood-optimization mechanism that links likelihood to correctness by sharpening the distribution toward sequences rewarded by the verification mechanism, while chain-of-thought makes final-answer likelihood reflect consistency with a sampled reasoning path rather than uncertainty about the answer itself.
Prior work estimates confidence with external signals or auxiliary modules: scoring rules built on model outputs \citep{hendrycks2017baseline,guo2017calibration}, probes on internal representations \citep{azaria2023internal,burns2023discovering}, or selector models trained alongside the answerer  \citep{geifman2019selectivenet,madras2018predict,mozannar2020consistent,whitehead2022reliable,srinivasan2024recovrr}.
More recently, LLMs themselves have been used for selective prediction by asking them to verbalize a confidence score alongside their answer \citep{lin2022teaching,kadavath2022language,tian2023just,xiong2023llms}.
Without explicit training, models verbalize confidence poorly, so reinforcement learning that explicitly targets confidence prediction has therefore become the leading approach to obtain more reliable responses from models \citep{damani2025beyond,xuan2026confidence,wu2026behaviorally}.
The rewards used to date score each confidence in isolation, pushing it toward $1$ when the answer is correct and toward $0$ when it is wrong \citep{damani2025beyond,xuan2026confidence,wu2026behaviorally}.
At inference, confidence is used to select only the most confident answers to improve task coverage at a chosen risk. 
What matters is therefore how each value ranks relative to that threshold rather than where it lands on the unit interval.
Risk tolerances vary across applications and evaluation settings (a coding orchestrator may accept a worker agent's lower-confidence patch when downstream tests will catch errors, whereas a VQA assistant for blind users or an autonomous vehicle can only act on high-confidence outputs), so the same confidence value can clear the abstention threshold under one operating point and fall short under another.
A reward that scores each confidence in isolation therefore reduces the supervision to a push toward $0$ or $1$, and in practice produces polarized confidences, weak low-risk coverage, and poor generalization of selective prediction.

We propose to train confidence as a cross-question signal for which each per-completion target is conditioned on the other answers given by the model to the same prompt and other prompts, rather than only on its own correctness label.
Selective prediction evaluates confidence over a collection of answers: for any chosen risk tolerance, a threshold determines which answers would be accepted and which would be abstained.
We therefore reward each confidence by the selection behavior it would induce across the set, rather than by a fixed correctness target.
Given a set of recent question-answer pairs with their confidence scores and correctness labels, we sort the set by confidence and identify, at a chosen risk tolerance, the largest subset of top-confidence answers whose error rate stays within tolerance.
The confidence at the boundary of this acceptable subset defines the abstention threshold for that risk level, and each completion is then rewarded according to where its own confidence sits relative to this threshold: lying above is reinforced for correct answers, and lying below is reinforced for incorrect ones.
Each risk tolerance induces its own threshold, so when reinforcing abstention at different risk levels, the same confidence value receives different rewards per risk level.
Because every threshold depends on the set, the supervision a model gets on any one answer reflects how that answer compares to its peers rather than any fixed correctness target.
Furthermore, we find that encouraging the model to predict diverse confidence scores, rather than collapsing around round numbers (e.g.\ single-decimal scores such as 0.8 or 0.9, or centesimals ending only in 0 or 5), improves coverage at low risk tolerances by preserving higher confidence-score resolution.
Notably, this signal can be added to the RLVR post-training phase in LLM training \citep{openai2024o1,deepseek2025r1} at no extra cost: no separate confidence model or training stage, only additional rewards in the existing RL objective.

We evaluate \ours in two training regimes: math (Big-Math-Digits) and multi-hop Wikipedia QA (HotpotQA).
\cameraReadyEdit{\ours substantially improves coverage at low risk, both within the training domain and on transfer to other domains, while maintaining competitive or even improving answer accuracy.
Compared with RLCR, coverage at 5\% risk rises from $0.8\%$ to $42.4\%$ on Math ID and from $12.3\%$ to $23.6\%$ on HotpotQA OOD.
These improvements extend across risk levels, yielding lower area under the risk-coverage curve (AURC).
}

\cameraReadyEdit{\paragraph{Contributions.}
We summarize our contributions as follows:
\begin{itemize}
\item We propose a cross-question coverage reward to train an LLM to verbalize confidence in its own answers, conditioning each completion's reward on abstention thresholds estimated from a set of answers rather than only its own binary correctness label, to maximize the number of questions answered within a user-specified risk tolerance.
\item We introduce rewards that encourage diverse confidence scores across answers, preventing collapse onto a small set of values and thereby preserving the resolution needed for fine-grained ranking and more precise abstention decisions at low risks.
\item We show that keeping confidence learning on-policy matters more than increasing the set of answers used to estimate the coverage reward: larger sets can reduce estimation noise, but accumulating answers from older rollouts makes the learning signal off-policy.
\end{itemize}
}

\section{Related Work}
\label{sec:related-work}

\paragraph{Selective prediction.}
Selective prediction tries to maximize the utility of models (e.g. their accuracy) under some risk requirements, by only solving the subset of tasks the model is most confident about \cite{chow1970survey}.
Approaches which directly learn to abstain \citep{cortes2016boosting,geifman2019selectivenet,xin2021artofabstention} cannot adapt to the user's risk tolerance, which might be more strict in high-stakes scenarios.
In contrast, a confidence score assigned to each answer can be used to satisfy various risk tolerances with a flexible abstention policy.
If available, the model's estimated likelihood of the chosen label is an intuitive candidate for such a confidence score.
For instance, for classification tasks, the predicted class' probability can be readily used to decide when to abstain \citep{geifmanSelectiveClassificationDeep2017}.
Recently, similar approaches have been extended to large language models \citep{yoshikawa-okazaki-2023-selective,chen-etal-2023-adaptation}.
However, due to their autoregressive nature, it is less clear which likelihood should represent certainty in the final answer.
The likelihood of a full chain-of-thought includes many intermediate tokens unrelated to the final answer, while the likelihood of the final tokens is heavily conditioned on the preceding reasoning trace \citep{nakkiran2025trained}.
Post-training phases which give rise to LLMs' best capabilities, like reinforcement learning with verifiable rewards (RLVR) \citep{openai2024o1,deepseek2025r1}, weaken likelihood as a confidence signal \citep{nakkiran2025trained}.
An alternative line of work uses an external selector to output confidence scores from the question, answer, supporting evidence, or internal representations such as embeddings or logits \citep{rodriguez2026sieves,whitehead2022reliable,dancette2024selective,mushtaq2025harmony,srinivasan2024recovrr}.
These approaches duplicate knowledge across the answerer and the selector, add inference cost, and often require internal representations that are unavailable for stronger proprietary models.
In contrast, we teach the answering model to reason about confidence in its own answer, simplifying inference to a single model that can already use its task knowledge.

\paragraph{Verbalized confidence prediction.}
Given the limitations of using an external selector or model intrinsics for confidence prediction, recent methods ask LLMs to verbalize confidence by outputting a numerical score in language tokens alongside each answer \citep{tian2023just, xiong2023llms}.
LLMs' zero-shot confidence prediction performance is limited, so \citet{damani2025beyond} propose learning confidence directly during the RLVR phase through an additional reward at no extra cost.
More concretely, RLCR \citep{damani2025beyond} trains LLMs' verbalized confidence with a Brier-style reward, equivalent to a mean-squared error between the confidence score and the binary correctness label, so each score is pushed toward 1 for a correct answer and 0 for an incorrect one.
A second line of work keeps this pointwise structure but reshapes the reward to protect accuracy.
\citet{xuan2026confidence} propose CAR (Calibration Agentic RL), which adds a margin-separated reward to RLCR's Brier reward that preserves a gap between correct and incorrect answers independent of predicted confidence, in order to avoid reward hacking.
\citet{wu2026behaviorally} (BC-Beta) recast RLCR's Brier reward as integrating abstention behavior uniformly over all possible risk thresholds, and introduce a prior over user risk tolerances using a symmetric Beta distribution that up-weights both very-low-risk (high-stakes) and very-high-risk (high-coverage) regimes.
These methods therefore target accuracy within the pointwise confidence-learning setting rather than changing the structure of the confidence signal itself.
In contrast, \ours optimizes confidence prediction using a set of predicted answer-confidence pairs, with a new reward that induces better coverage at risk rather than matching individual correctness.
Because the two axes are complementary, accuracy-oriented refinements such as CAR's margin term or BC-Beta's risk prior could in principle be combined with our cross-question coverage rewards.

\paragraph{Learning from other samples.}
Several methods exploit information from other samples during learning: transductive learning \citep{vapnik1998statistical,joachims1999transductive}, test-time adaptation \citep{hardt2024testtime,zhang2025scap}, and learning to rank, where pairs or lists of examples provide supervision for their relative ordering \citep{burges2005learning,cao2007learning}.
\cameraReadyEdit{Concurrently to our work, RLSR \citep{luo2026aligning} and ORCE \citep{li2026orce} propose using other completions to compute confidence-related rewards.
RLSR optimizes an AURC-based reward that depends only on confidence rankings, unlike \ours, which also uses confidence values to provide a stronger signal.
ORCE trains a separate confidence model using DPO, with preferences determined by each prediction's contribution to the correlation between confidence rankings and correctness rates.
Measuring an individual prediction's contribution to this set-wide statistic dilutes the learning signal.
In contrast, \ours introduces a confidence reward that has a much stronger dependence on the correctness and confidence prediction of the completion being rewarded, while still being conditioned on the cross-question set of completions.
}

\section{\oursLong (\ours)}
\label{sec:method}

Obtaining sensible confidence scores associated with a model's answers, and hence introducing a mechanism to abstain or defer to a human or model expert, can help maximize the share of questions answered automatically while keeping the risk of incorrect answers low.
However, the ability of LLMs to do this zero-shot is limited, both through their intrinsic likelihoods or through verbalized confidence, especially when combined with chain-of-thought reasoning and post-training~\citep{nakkiran2025trained}, see \Cref{sec:related-work}.
Recent methods therefore train LLMs to express the confidence in their own answer outputting scores using standard tokens during the RLVR phase~\citep{damani2025beyond,xuan2026confidence,wu2026behaviorally}.
However, existing objectives reward confidence against binary correctness labels, rather than directly targeting coverage at the user's risk tolerance.
\ours, in contrast, optimizes confidence prediction over a set of answers and confidence scores, enhancing their relative scores rather than optimizing each toward a binary target.
Specifically, a \emph{coverage reward} (\cref{sec:method-coverage}) computes the abstention threshold from a set of question-answer pairs at a chosen risk tolerance and reinforces each confidence on the selection-preserving side.
A \emph{diversity-inducing reward} (\cref{sec:method-reg}) mechanism encourages the model to predict fine-grained confidence values and prevents collapse to coarse scores that are more natural in language like using single digits (e.g.\ 0.8 or 0.9) or multiples of 5 (e.g.\ 0.15 or 0.10).

\begin{figure}[t]
    \centering
    \includegraphics[width=\linewidth]{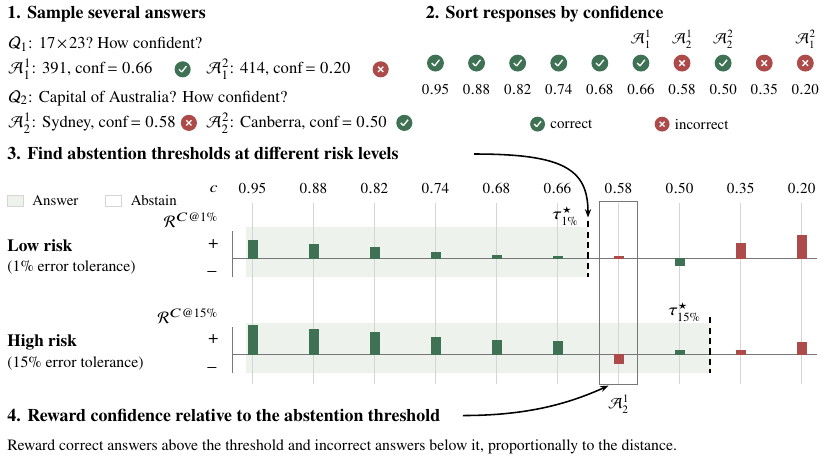}
    \caption{\cameraReadyEdit{\textbf{\ours rewards confidence relative to abstention thresholds estimated across sets of questions and answers.} \textbf{1.} Sample several answers and confidence scores for each question. \textbf{2.} Pool the answers across questions and sort them by confidence. \textbf{3.} For several error tolerances, find the confidence thresholds that maximize coverage while keeping the accepted answers' error rate within tolerance. \textbf{4.} Reward correct answers above the threshold and incorrect answers below it, and penalize the opposite placements, with reward magnitude increasing with distance from the threshold. The highlighted incorrect answer illustrates how the same confidence score can receive a positive reward under a stricter error tolerance and a negative reward under a more permissive one.}}
    \label{fig:care-method}
\end{figure}

\cameraReadyEdit{\subsection{Reinforcing coverage at a chosen risk tolerance}
\label{sec:method-coverage}
We first define coverage at risk, then use its confidence threshold to construct the training reward.

\paragraph{Coverage at risk (C@r).}
Given $N$ answers, each with confidence score $c_i$ and correctness label $y_i$, coverage at risk (C@r) measures the largest fraction that can be accepted while keeping its empirical error rate at most $r$.
For an abstention threshold $\tau$, coverage and selective error rate are defined as
\begin{equation}
C(\tau) = \frac{1}{N}\sum_{i=1}^{N}\indicator[c_i \ge \tau],
\qquad
R(\tau) = \frac{\sum_{i=1}^{N}(1 - y_i)\,\indicator[c_i \ge \tau]}{\sum_{i=1}^{N}\indicator[c_i \ge \tau]},
\end{equation}
where $y_i\in\{0,1\}$ indicates whether the answer is correct.
We define coverage at user-defined risk $r$ as the maximum coverage attained while the selective error rate $R(\tau)$ does not surpass the user's risk tolerance $r$.
More precisely,
\begin{equation}
\mathrm{C@}r=\max_{\tau:\,R(\tau)\le r}C(\tau),\qquad
\tau^\star(r)\in\argmax_{\tau:\,R(\tau)\le r}C(\tau).
\label{eq:care-constrained-threshold}
\end{equation}
}

\paragraph{Coverage reward formulation.}
At training, we introduce an objective that aligns with C@r.
Using a set of question-answer pairs $\mathcal{S} = \{(c_i, y_i)\}_{i=1}^{N}$, represented by their confidence scores and correctness labels,
\cameraReadyEdit{we estimate the risk-constrained threshold $\tau^\star(r)$.
We then reward each completion according to where its own confidence sits relative to this threshold (\Cref{fig:care-method}).
Concretely, 
\begin{equation}
\mathcal{R}^{\text{C@}r} = (2y - 1)\operatorname{sign}\!\left(c-\tau^\star(r)\right)\bigl|c - \tau^\star(r)\bigr|^p,
\end{equation}
so correct answers receive positive reward above $\tau^\star(r)$ and negative reward below it, while incorrect answers receive the opposite signs.
At the threshold the reward is zero.
For example, at $\tau^\star=0.5$ and $p=2$, a correct answer at $c=0.7$ receives $+0.04$, while one at $c=0.3$ receives $-0.04$.
}
$p$ is a hyperparameter which highlights scores closer or further away from the abstention threshold, see \Cref{tab:reward-geometry-ablation-master}.
Thus this reward does not reinforce whether $c$ matches the answer's correctness label in isolation, but whether $c$ places the completion on the correct side of the abstention threshold that optimizes coverage at risk $r$.
Because $\tau^\star(r)$ is recomputed from the model's current set of answers, the threshold reflects the model's achievable coverage at that point in training.
Training over multiple risk tolerances exposes the same confidence distribution to several abstention thresholds, encouraging scores that support selection across potential downstream risk tolerances rather than only at one threshold, and hence providing a more fine-grained training signal.
The size and scope of $\mathcal{S}$ trade off on-policy estimates against more robust abstention thresholds, as studied in \Cref{sec:results-ablations}.

\subsection{Reinforcing diverse confidence scores}
\label{sec:method-reg}
\providecommand{\graycell}[1]{\textcolor{gray}{#1}}

\begin{table*}[t]
\centering
\caption{\cameraReadyEdit{\textbf{Selective prediction training on math (in \%):} ID (Big-Math-Digits, MATH-500, GSM8K), OOD (general knowledge \& reasoning: CommonsenseQA, GPQA, SimpleQA, TriviaQA, HotpotQA).
\textcolor{gray}{RLVR-Classifier} is the only method not using verbalized confidence prediction from an LLM.
\textbf{Bold} and \underline{underlining} mark the best and second-best results among methods with verbalized confidence.}}
\label{tab:seed1-math-id-ood-raw}
\small
\setlength{\tabcolsep}{4pt}
\resizebox{\textwidth}{!}{\begin{tabular}{llcccc>{\color{gray}}ccccc>{\color{gray}}c}
\toprule
\multicolumn{2}{c}{} & \multicolumn{5}{c}{\textbf{ID}} & \multicolumn{5}{c}{\textbf{OOD}} \\
\cmidrule(lr){3-7}\cmidrule(lr){8-12}
\textbf{Model} & \textbf{Init} & \textbf{C@0} & \textbf{C@1} & \textbf{C@5} & \textbf{AURC\,$\downarrow$} & \textbf{Acc} & \textbf{C@0} & \textbf{C@1} & \textbf{C@5} & \textbf{AURC\,$\downarrow$} & \textbf{Acc} \\
\midrule
\graycell{RLVR-Classifier \citep{damani2025beyond}} & \graycell{N/A} & \graycell{4.6} & \graycell{5.9} & \graycell{33.6} & \graycell{12.8} & \graycell{73.7} & \graycell{0.5} & \graycell{0.5} & \graycell{1.5} & \graycell{45.2} & \graycell{52.5} \\
\midrule
CAR-WeightedBrier \citep{xuan2026confidence} & Cold & 0.0 & 0.0 & 0.0 & 20.6 & 71.8 & \underline{0.1} & \underline{0.1} & 0.1 & 41.7 & 51.3 \\
CAR-MSCR \citep{xuan2026confidence} & Cold & 0.0 & 0.0 & 0.0 & 19.2 & 72.0 & \underline{0.1} & \underline{0.1} & 7.0 & \underline{37.7} & \underline{56.7} \\
BC-Beta \citep{wu2026behaviorally} & Cold & 0.0 & 0.0 & 0.0 & 17.6 & 71.8 & 0.0 & 0.0 & \underline{11.6} & \textbf{37.6} & 55.7 \\
\midrule
RLSR \citep{luo2026aligning} & Cold & 0.0 & 0.0 & 0.0 & 21.7 & 72.7 & 0.0 & 0.0 & 0.0 & 51.0 & 43.9 \\
ORCE-SC \citep{li2026orce} & Cold & 0.0 & 0.0 & 0.0 & 23.6 & \textbf{75.0} & 0.0 & 0.0 & 0.0 & 40.9 & \textbf{56.8} \\
\midrule
RLVR \citep{damani2025beyond} & Cold & 0.0 & 0.0 & 0.0 & 31.0 & 70.3 & 0.0 & 0.0 & 0.0 & 50.1 & 49.5 \\
RLCR \citep{damani2025beyond} & SFT & 0.0 & 0.0 & 0.0 & \underline{16.3} & 72.0 & 0.0 & 0.0 & 0.0 & 42.0 & 50.0 \\
RLCR \citep{damani2025beyond} & Cold & 0.0 & 0.0 & \underline{0.8} & 19.3 & \underline{73.1} & 0.0 & 0.0 & 4.9 & 38.7 & 56.6 \\
\textbf{CARE (Ours)} & Cold & \textbf{5.4} & \textbf{18.2} & \textbf{42.4} & \textbf{12.5} & \underline{73.1} & \textbf{2.9} & \textbf{5.4} & \textbf{15.9} & \underline{37.7} & 53.6 \\
\bottomrule
\end{tabular}}
\end{table*}

To enhance correct relative ordering of answers based on confidence, we want the model to use varied confidence scores.
Concretely, coverage at low risk depends on fine distinctions among high-confidence answers that are not represented if completions collapse onto coarse scores such as 0.8 or 0.9.
We therefore add two diversity-inducing rewards, one operating on the coarse shape of the confidence histogram and one on individual values.
We study their contribution in \Cref{sec:results-ablations}.

\paragraph{Entropy reward.}
This reward encourages confidence mass to spread across coarse regions of $[0,1]$.
We partition $[0,1]$ into $B$ equal-width bins, where bin $j$ covers $[(j-1)/B, j/B)$ for $j < B$, and the final bin includes $1$.
Let $j(c)$ denote the index of the bin containing confidence $c$.
The normalized mass of bin $j$ over the set $\mathcal{S}$ is
\begin{equation}
m_j = \frac{1}{|\mathcal{S}|}\sum_{(c_i,y_i)\in\mathcal{S}}\indicator[j(c_i) = j].
\end{equation}
The uniform target mass is $1/B$.
For a completion with confidence $c$, the raw reward is
\begin{equation}
\tilde{\mathcal{R}}^{\text{ent}}(c) = 1 - B\,m_{j(c)}.
\end{equation}
Thus underused bins receive positive reward, uniformly occupied bins receive zero reward, and overused bins receive negative reward.
We clip the reward below at $-1$ to keep highly crowded bins from dominating, giving the entropy signal
\begin{equation}
\mathcal{R}^{\text{ent}}(c) = \max\{-1,\tilde{\mathcal{R}}^{\text{ent}}(c)\}.
\end{equation}

\paragraph{Uniqueness reward.}
Whereas the entropy reward operates on coarse bins, models often default to a small set of specific confidence values.
The uniqueness reward therefore acts directly on raw values, encouraging distinct confidence predictions within a set $\mathcal{S}$.
With tolerance $\delta$,
\begin{equation}
n^{\text{sim}}(c) = \sum_{(c_i,y_i)\in\mathcal{S}} \indicator\!\left[|c_i - c| \le \delta\right],
\qquad
\mathcal{R}^{\text{uniq}}(c) = 1 - 2\frac{n^{\text{sim}}(c)}{|\mathcal{S}|}.
\end{equation}
Thus maximum $\mathcal{R}^{\text{uniq}}$ is assigned when the predicted confidence is unique in $\mathcal{S}$, with proportional reduction as more of the set repeats the same value.

\subsection{Final objective}

Building on existing confidence-prediction frameworks~\citep{damani2025beyond}, we combine the coverage and diversity rewards with three per-completion signals carried over from prior work: a binary accuracy reward $\mathcal{R}^{\text{ans}}$ indicating whether the answer is correct, a format reward $\mathcal{R}^{\text{fmt}}$ that rewards emitting the answer and confidence in the expected output format, and a pointwise Brier confidence reward $\mathcal{R}^{\text{brier}} = 1 - (y - c)^2$:
\cameraReadyEdit{\begin{align}
\mathcal{R} ={}&
w_{\text{fmt}}\,\mathcal{R}^{\text{fmt}} +
w_{\text{ans}}\,\mathcal{R}^{\text{ans}} +
w_{\text{brier}}\,\mathcal{R}^{\text{brier}} \nonumber\\
&+
\sum_{r \in \mathcal{T}} w_{\text{C@},r}\,\mathcal{R}^{\text{C@}r} +
w_{\text{uniq}}\,\mathcal{R}^{\text{uniq}} +
w_{\text{ent}}\,\mathcal{R}^{\text{ent}},
\end{align}
where $\mathcal{T}$ is the set of risk tolerances included in training, and $w$ represents the scalar weights, see \Cref{app:experimental-details}.
We maintain the Brier reward for consistency with existing methods, and as a possible source of stable confidence-prediction learning in early stages of training, when coverage rewards can be noisier (\Cref{app:rebuttal-brier}).
In practice, we find that removing it has little or even positive impact, see \Cref{sec:results-ablations}.
}

\section{Experiments}
\label{sec:experiments}

\providecommand{\graycell}[1]{\textcolor{gray}{#1}}

\paragraph{Training and benchmarks.} 
We follow the training and evaluation protocol of \citet{damani2025beyond} (RLCR), and consider two separate training settings: a \emph{math} setting that trains on Big-Math-Digits~\citep{albalak2025big}, and a \emph{knowledge} setting that trains on HotpotQA~\citep{yang2018hotpotqa}.
In each setting, we train Qwen2.5-7B~\citep{yang2024qwen25} with GRPO~\citep{shao2024deepseekmath} and evaluate transfer to a suite of unseen benchmarks (Appendix Table~\ref{tab:datasets}).
For math, these are MATH-500~\citep{hendrycks2021math} and GSM8K~\citep{cobbe2021training}, and CommonsenseQA~\citep{talmor2019commonsenseqa}, GPQA~\citep{rein2023gpqa}, SimpleQA~\citep{wei2024simpleqa}, TriviaQA~\citep{joshi2017triviaqa}, and HotpotQA for general knowledge and reasoning.
We also follow the same evaluation protocol as \citet{damani2025beyond}.
\citet{damani2025beyond} use a short SFT warm start before RL on Big-Math-Digits, but not HotpotQA.
We find that cold-start matches or exceeds it (see \Cref{tab:seed1-math-id-ood-raw}) and do not perform SFT for \ours or other baselines.
Further details are in \Cref{app:experimental-details}.

\paragraph{Selective prediction metrics.} The question is whether confidence is \emph{useful} for selective prediction: how many inputs can the system answer while keeping the error rate among answered inputs bounded?
We therefore report \textbf{coverage at risk} (C@r) \citep{geifmanSelectiveClassificationDeep2017,geifman2019selectivenet} and \textbf{area under the risk--coverage curve} (AURC) \citep{geifman2019biasreduced}, for a more global view of coverage.
Standard calibration metrics such as ECE or Brier score do not directly answer this question: for example, a selector emitting a constant near the base accuracy already attains acceptable ECE while not enabling selection of some answers over others.
Nonetheless, we include a calibration study in \Cref{app:rebuttal-calibration}.

\paragraph{Baselines.} We compare against eight baselines.
\textbf{RLVR} \citep{damani2025beyond} takes a model trained with GRPO on accuracy alone and prompts it zero-shot to verbalize a confidence score. 
\textbf{RLVR-Classifier} \citep{damani2025beyond} is a separately finetuned selector that predicts answer correctness from RLVR generations.
This is not entirely comparable to the other settings, as the capacity for answering and confidence prediction is now duplicated, and separate optimization focused only on confidence prediction can be performed.
Similarly, at inference, this method is twice as expensive as both answering and confidence prediction models (i.e.\ classifier) must be run.
\textbf{RLCR} \citep{damani2025beyond} adds a pointwise Brier-style confidence reward to GRPO and is the closest direct comparison to \ours.
\textbf{CAR-WeightedBrier} \citep{xuan2026confidence}, \textbf{CAR-MSCR} \citep{xuan2026confidence}, and \textbf{BC-Beta} \citep{wu2026behaviorally} are recent variants of the pointwise confidence reward that aim to balance accuracy and coverage: 
CAR adds a reward gap between correct and incorrect answers, and BC-Beta places a prior over risk tolerances that up-weights the very-low-risk and very-high-risk regimes.
We use the default hyperparameters ($\lambda = 1/3$, $\beta_1 = \beta_2 = 1$, and a truncated $\mathrm{Beta}(0,0)$ prior, respectively).
\cameraReadyEdit{Concurrently to our work, {RLSR} \citep{luo2026aligning} and ORCE \citep{li2026orce} introduce confidence rewards computed from multiple completions.
\textbf{RLSR} rewards correct answers and penalizes incorrect ones according to their confidence ranks using an AURC-based objective.
ORCE rewards confidence predictions that improve the ranking of questions with higher success rate.
We adapt these reward designs to our setup, denoting the ORCE adaptation as \textbf{ORCE-SC}.
\Cref{app:experimental-details} contains implementation details.
}

\section{Results}
\label{sec:results}

\providecommand{\graycell}[1]{\textcolor{gray}{#1}}

Following \cite{damani2025beyond}, we report results for both training regimes math and HotpotQA, on in- and out-of-distribution benchmarks.
Other risk levels and additional metrics are reported in \Cref{app:rebuttal-calibration}.

\subsection{Main results}
\label{sec:results-main}

\providecommand{\graycell}[1]{\textcolor{gray}{#1}}

\begin{table*}[t]
\centering
\caption{\cameraReadyEdit{\textbf{Selective prediction training on HotpotQA:} ID (HotpotQA) \& OOD (General Knowledge, Reasoning, Math: MATH-500, GSM8K, CommonsenseQA, GPQA, SimpleQA, TriviaQA).
\textcolor{gray}{RLVR-Classifier} is the only method not using verbalized confidence prediction from an LLM.
\textbf{Bold} and \underline{underlining} mark the best and second-best results among methods with verbalized confidence.}}
\label{tab:seed1-hotpot-id-ood-raw}
\small
\setlength{\tabcolsep}{4pt}
\resizebox{\textwidth}{!}{\begin{tabular}{lcccc>{\color{gray}}ccccc>{\color{gray}}c}
\toprule
\multicolumn{1}{c}{} & \multicolumn{5}{c}{\textbf{ID}} & \multicolumn{5}{c}{\textbf{OOD}} \\
\cmidrule(lr){2-6}\cmidrule(lr){7-11}
\textbf{Model} & \textbf{C@0} & \textbf{C@1} & \textbf{C@5} & \textbf{AURC\,$\downarrow$} & \textbf{Acc} & \textbf{C@0} & \textbf{C@1} & \textbf{C@5} & \textbf{AURC\,$\downarrow$} & \textbf{Acc} \\
\midrule
\graycell{RLVR-Classifier \citep{damani2025beyond}} & \graycell{1.6} & \graycell{1.6} & \graycell{14.2} & \graycell{6.5} & \graycell{87.7} & \graycell{0.4} & \graycell{0.4} & \graycell{6.0} & \graycell{42.0} & \graycell{53.9} \\
\midrule
CAR-WeightedBrier \citep{xuan2026confidence} & 0.0 & 0.0 & 0.0 & 6.4 & 85.8 & 0.0 & 0.0 & \underline{16.5} & 35.6 & 57.3 \\
CAR-MSCR \citep{xuan2026confidence} & 0.0 & 0.0 & 0.0 & 7.2 & 87.0 & 0.0 & \textbf{7.4} & 14.8 & 36.2 & 56.0 \\
BC-Beta \citep{wu2026behaviorally} & 0.1 & 0.1 & 0.1 & 7.2 & 81.9 & \underline{0.9} & 5.6 & 14.8 & \textbf{30.5} & \textbf{63.4} \\
\midrule
RLSR \citep{luo2026aligning} & 0.0 & 0.0 & 0.0 & 8.6 & 86.6 & 0.0 & 0.0 & 14.5 & 37.7 & \underline{58.8} \\
ORCE-SC \citep{li2026orce} & 0.0 & 0.0 & 0.0 & 10.7 & 85.6 & 0.0 & 0.0 & 14.9 & 38.5 & 55.2 \\
\midrule
RLVR \citep{damani2025beyond} & 0.0 & 0.0 & 0.0 & 12.3 & \textbf{87.7} & 0.0 & 0.0 & 0.0 & 46.0 & 53.9 \\
RLCR \citep{damani2025beyond} & \underline{2.4} & \underline{2.4} & \underline{34.7} & \underline{6.1} & 86.1 & 0.1 & 0.1 & 12.3 & 36.6 & 56.2 \\
\textbf{CARE (Ours)} & \textbf{6.1} & \textbf{6.1} & \textbf{74.7} & \textbf{4.4} & \underline{87.6} & \textbf{3.1} & \underline{6.4} & \textbf{23.6} & \underline{33.2} & 58.0 \\
\bottomrule
\end{tabular}}
\end{table*}

\begin{figure*}[t]
    \centering
    \includegraphics[width=\textwidth]{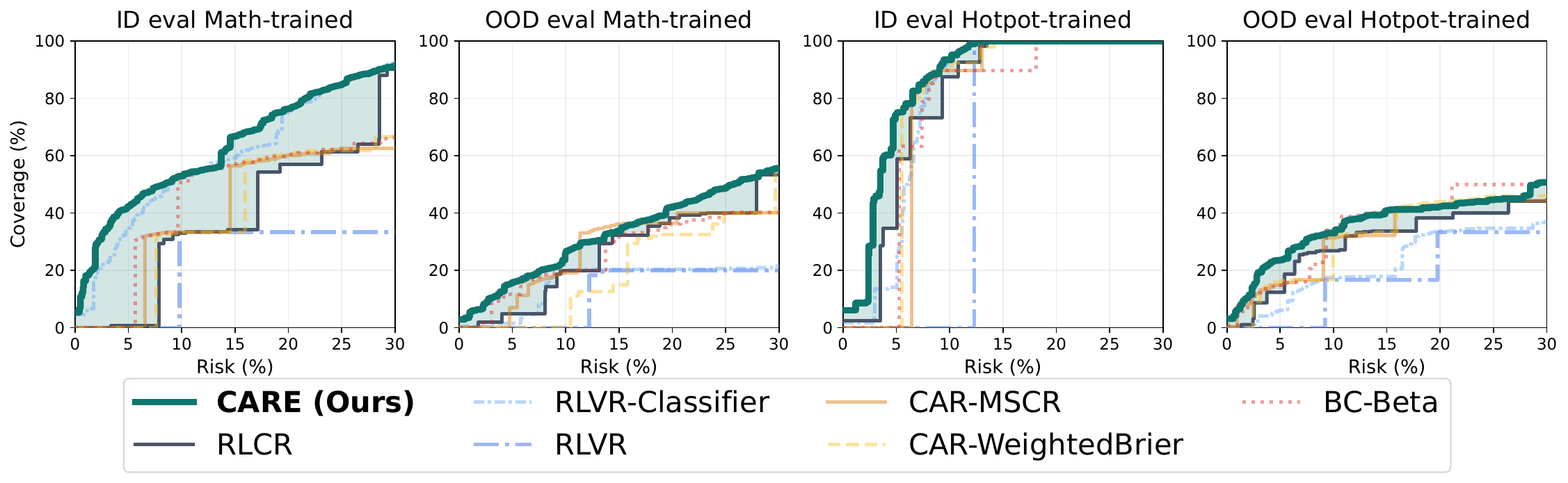}
    \caption{
        \textbf{\ours achieves higher coverage across most risk levels and tasks} on in-distribution (ID) and out-of-distribution (OOD) benchmarks, under both training regimes.
        \ours attains consistently higher coverage, with much smoother curves that allow coverage at any risk tolerance.
        Shading highlights the difference to RLCR, our most direct comparison.
    }
    \label{fig:seed1-id-ood-risk-coverage}
\end{figure*}

\paragraph{\ours unlocks low-risk coverage.}
Across both training regimes and evaluation settings (\Cref{tab:seed1-math-id-ood-raw,tab:seed1-hotpot-id-ood-raw}), \ours consistently unlocks coverage at low risk:
C@0\%, C@1\%, and C@5\% generally rise by a wide margin over RLVR, RLCR, as well as CAR and BC-Beta variants, and even the RLVR-Classifier.
As can be seen in \cref{fig:seed1-id-ood-risk-coverage}, \ours provides much better coverage at low risks.
This is the failure mode of pointwise training: a model can look calibrated on average while still failing to order its highest-confidence answers sharply enough for low-risk abstention.
\cameraReadyEdit{Furthermore, \ours improves coverage over RLSR and ORCE-SC, likely due to the more direct per-completion feedback provided by \ours.
RLSR uses only the relative ranking between confidence scores, not numerical confidence differences as \ours does, providing sparser feedback.
ORCE-SC measures each prediction's marginal contribution to a set-wide correlation, which dilutes the individual learning signal.
In contrast, \ours provides more guidance, rewarding each completion based on the difference between its own confidence and the abstention threshold computed from the set of completions.
}

\paragraph{\ours improves coverage beyond highest stakes.}
\cref{fig:seed1-id-ood-risk-coverage} shows that, even though we optimize specific risk levels during training, the coverage curve progresses smoothly with increased risk, unlike existing methods.
\cameraReadyEdit{\ours improves AURC over RLCR in all training and evaluation scenarios, demonstrating that the improvements in coverage are not limited to the highest stakes.
The only case where an alternative method can compete with \ours in terms of AURC is OOD evaluation, where BC-Beta achieves higher accuracy and therefore provides better AURC.
However, its confidence prediction is worse at low risk tolerance (\Cref{fig:seed1-id-ood-risk-coverage}) and ID performance is worse overall, including accuracy (e.g.\ 87.6\% for \ours on ID HotpotQA vs.\ 81.9\% for BC-Beta).
}
Combining an accuracy-maximizing objective (orthogonal to our contribution) with \ours could recover such behavior.

\paragraph{Accuracy is preserved or improved.}
Compared to its closest comparison, RLCR, \ours matches or improves accuracy across most splits, and substantially improves over RLVR, which trains with accuracy alone, on both ID and OOD averages.
Adding cross-question coverage and diversity rewards, therefore, does not trade off task accuracy for confidence quality.
CAR-WeightedBrier and BC-Beta target a different axis: they sharpen accuracy within the confidence-learning setting via strict-margin rewards that preserve a gap between correct and incorrect answers (CAR) or via priors over risk tolerances that emphasize the very-low-risk and very-high-risk regimes (BC-Beta).
They sometimes attain higher accuracy than \ours, especially on some OOD non-math tasks, but generally remain inferior at strict risk thresholds.
Their accuracy refinements act on a complementary axis to our cross-question coverage rewards and could, in principle, be combined with \ours.

\begin{figure*}[t]
\centering
\includegraphics[width=\textwidth]{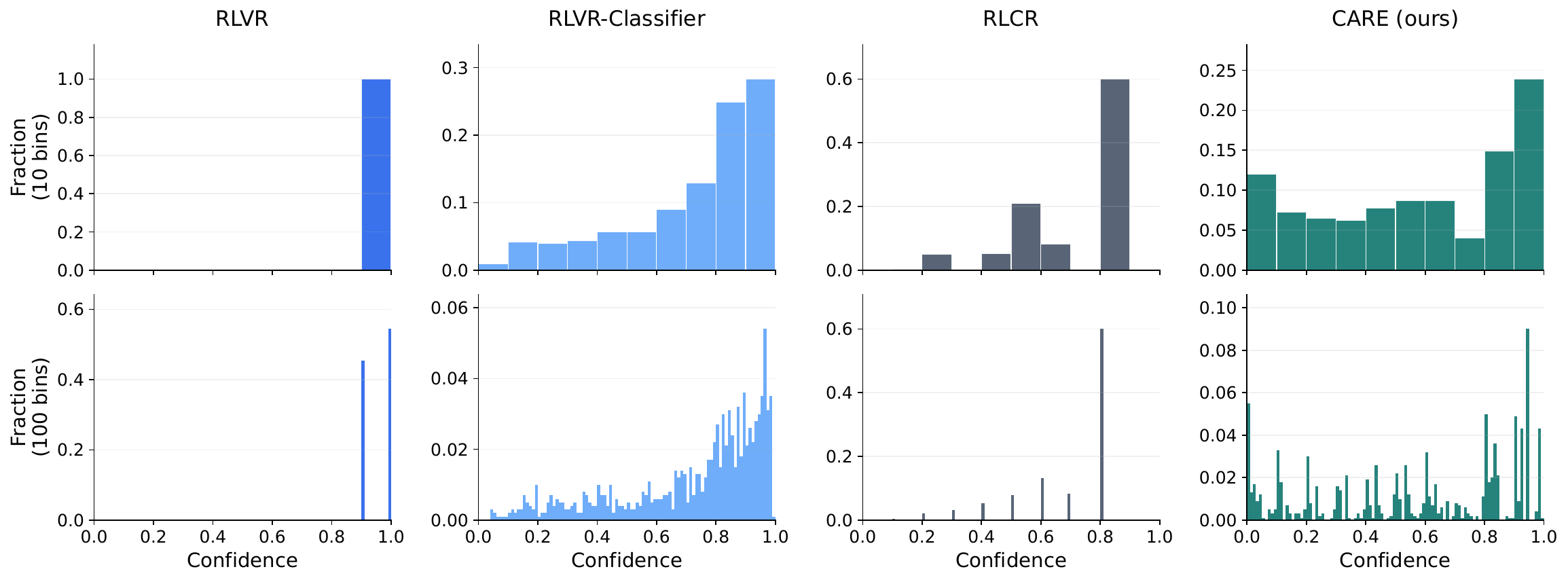}
\caption{
    \textbf{Histogram of predicted confidence values} on Big-Math-Digits using 10 (top) and 100 (bottom) bins.
    Baselines are overconfident. 
    Verbalized baselines collapse predictions to simple decimals.
    \textbf{\ours predicts fine-grained confidence scores, enabling coverage at very low risks; and assigning low confidence for incorrect answers, allowing abstention for difficult tasks.}
}
\label{fig:math-confidence-distributions}
\end{figure*}

\Cref{fig:math-confidence-distributions} shows histograms of predicted confidence per model, grouped on two coarseness levels.
When looking at the coarser plots (top), we can see that the baselines (RLVR, RLVR-Classifier, and RLCR) concentrate on the high-confidence regime, hence often being over-confident, especially in the zero-shot RLVR case.
\ours, in contrast, frequently predicts low confidence for answers it is unsure about.
We can also look at the more fine-grained histogram at the bottom of \Cref{fig:math-confidence-distributions}.
When looking at the verbalized confidence prediction baselines, we observe that only single-digit values are used to predict confidence.
Even though confidence optimization introduced by RLCR optimizes the model away from the degenerate overconfidence of vanilla RLVR, we see that it does not induce diverse confidence scores.
In \Cref{app:additional-analyses} we show how the other brier-like confidence optimization baselines also follow this pattern. 
Only the RLVR-Classifier, which utilizes a linear head and hence a more natural continuous output space, and \ours utilize a wider range of scores.
\ours still frequently predicts round numbers with fewer decimals, but also uses many other values.

\cameraReadyEdit{\paragraph{Results on another model family and size.}
We also train Gemma 4 E2B-it~\citep{gemmateam2026gemma4}, a smaller model with 2.3B effective parameters from a different family than Qwen2.5-7B, in both training regimes.
Compared with RLCR, the closest baseline, \ours again unlocks in-distribution coverage at low risk where RLCR has none, and improves AURC and accuracy in both regimes, ID and OOD (\Cref{tab:rebuttal-gemma}; details in \Cref{app:gemma-details}), confirming that its benefits transfer across model families and scales.
}

\begin{table}[t]
\centering\footnotesize
\setlength{\tabcolsep}{2.5pt}
\caption{\cameraReadyEdit{\textbf{\ours improves confidence prediction for Gemma 4 E2B-it in both training and evaluation settings, demonstrating that its benefits generalize to other model families and sizes.}
Training on math or HotpotQA, with ID and OOD benchmarks as in \Cref{tab:seed1-math-id-ood-raw,tab:seed1-hotpot-id-ood-raw}.
All metrics are percentages. \textbf{Bold} marks the best result per regime.}}
\label{tab:rebuttal-gemma}
\begin{tabular*}{\linewidth}{@{\extracolsep{\fill}}l@{\hskip 10pt}lrrrrr>{\color{gray}}rrrrrr>{\color{gray}}r@{}}
\toprule
 & & \multicolumn{6}{c}{\textbf{ID}} & \multicolumn{6}{c}{\textbf{OOD}} \\
\cmidrule(lr){3-8}\cmidrule(lr){9-14}
\textbf{Train Data} & \textbf{Method} & \textbf{C@0} & \textbf{C@1} & \textbf{C@5} & \textbf{C@10} & \textbf{AURC$\downarrow$} & \textbf{Acc} & \textbf{C@0} & \textbf{C@1} & \textbf{C@5} & \textbf{C@10} & \textbf{AURC$\downarrow$} & \textbf{Acc} \\
\midrule
\multirow{2}{*}{Math} & RLCR \citep{damani2025beyond} & 0.0 & 0.0 & 0.0 & 0.0 & 18.2 & 73.9 & 0.0 & 0.0 & 4.9 & \textbf{33.1} & 38.0 & 54.0 \\
 & \textbf{\ours (Ours)} & \textbf{0.3} & \textbf{0.3} & \textbf{0.7} & \textbf{37.7} & \textbf{15.5} & \textbf{74.4} & \textbf{0.8} & \textbf{0.8} & \textbf{8.8} & 26.2 & \textbf{36.5} & \textbf{54.5} \\
\midrule
\multirow{2}{*}{HotpotQA} & RLCR \citep{damani2025beyond} & 0.0 & 0.0 & 0.0 & 92.3 & 6.8 & 86.7 & 0.0 & 0.0 & \textbf{7.8} & 14.1 & 36.1 & 55.7 \\
 & \textbf{\ours (Ours)} & 0.0 & 0.0 & \textbf{58.2} & \textbf{93.6} & \textbf{6.1} & \textbf{88.0} & \textbf{1.4} & \textbf{1.9} & 5.0 & \textbf{18.5} & \textbf{33.9} & \textbf{56.5} \\
\bottomrule
\end{tabular*}
\end{table}

\subsection{Ablations}
\label{sec:results-ablations}
\label{sec:results-reward-design}
\label{sec:results-buffer}

\paragraph{Reward design ablation.}

\Cref{tab:reward-design-ablation-id-ood} ablates the reward terms in our final objective when training in the math setting.
We observe that our design is robust to removing Brier reward (row~2 vs. row~1), which causes little or no harm, obtaining slightly better AURC and coverage at some risks.
We observe similar robustness when optimizing only the strictest risk level, C@0, instead of multiple risk levels (row~3 vs. row~1).
This is overall slightly worse, but generally better than existing methods.
However, removing our coverage rewards (row~4) is significantly damaging, highlighting the importance of our across-question coverage reward for good selective prediction.
This shows diversity encouraging rewards are not enough, brier reward alone does not encourage correct relative confidence scores. 
Dropping $\mathcal{R}^{uniq}$ (row~5) erodes C@5 (42.4 $\to$ 30.2), dropping $\mathcal{R}^{ent}$ (row~6) erodes it further (42.4 $\to$ 20.1).
Hence, diversity rewards are helpful and not redundant: the coarse-bin entropy term and the value-level uniqueness term each help enhance the resolution of confidence scores.

\begin{table*}[t]
\centering
\caption{\textbf{Reward-design ablations} in the math training regime. Row~1 (\#1) is the full \ours recipe. Removing Brier (\#2) largely preserves and even improves performance, whereas removing the coverage reward introduced by \ours (\#4) hurts performance the most.
Diversity rewards are useful and complementary (\#5,\#6), but not enough without the coverage reward (\#4).
Optimizing coverage at a single risk level (\#3) is worse than at multiple (\#1), but better than the baselines (\#4).
}
\label{tab:reward-design-ablation-id-ood}
\small
\setlength{\tabcolsep}{4pt}
\resizebox{\textwidth}{!}{\begin{tabular}{ccccccrrrrr>{\color{gray}}rrrrrr>{\color{gray}}r}
\toprule
\multicolumn{1}{c}{} & \multicolumn{5}{c}{\textbf{Reward terms}} & \multicolumn{6}{c}{\textbf{ID}} & \multicolumn{6}{c}{\textbf{OOD}} \\
\cmidrule(lr){2-6}\cmidrule(lr){7-12}\cmidrule(lr){13-18}
\textbf{\#} & $\mathcal{R}^{\text{brier}}$ & $\mathcal{R}^{\text{C@}0}$ & $\mathcal{R}^{\text{C@}r}$ & $\mathcal{R}^{\text{uniq}}$ & $\mathcal{R}^{\text{ent}}$ & \textbf{C@0} & \textbf{C@1} & \textbf{C@5} & \textbf{C@10} & \textbf{AURC$\downarrow$} & \textbf{Acc} & \textbf{C@0} & \textbf{C@1} & \textbf{C@5} & \textbf{C@10} & \textbf{AURC$\downarrow$} & \textbf{Acc} \\
\midrule
1 & \checkmark & \checkmark & \checkmark & \checkmark & \checkmark & \textbf{5.4} & \textbf{18.2} & \textbf{42.4} & 52.9 & \underline{12.5} & 73.1 & \textbf{2.9} & \underline{5.4} & 15.9 & 26.4 & 37.7 & 53.6 \\
2 & $\times$ & \checkmark & \checkmark & \checkmark & \checkmark & 3.0 & 3.0 & \underline{40.3} & \textbf{55.8} & \textbf{12.1} & \textbf{73.9} & \underline{2.6} & 4.0 & \underline{16.1} & 22.1 & 36.9 & 56.5 \\
3 & \checkmark & \checkmark & $\times$ & \checkmark & \checkmark & 2.2 & 3.0 & 33.1 & \underline{54.2} & 12.7 & 72.3 & 0.5 & 2.5 & 14.4 & \underline{28.7} & 36.8 & 53.7 \\
4 & \checkmark & $\times$ & $\times$ & \checkmark & \checkmark & 0.6 & 0.6 & 16.3 & 33.3 & 19.3 & 72.4 & 0.4 & 0.4 & 4.6 & 20.9 & 41.5 & 55.7 \\
5 & \checkmark & \checkmark & \checkmark & $\times$ & \checkmark & \underline{3.6} & \underline{3.6} & 30.2 & 51.6 & 13.2 & 72.3 & 2.0 & \textbf{6.9} & \textbf{18.5} & 26.3 & \textbf{33.5} & \underline{59.7} \\
6 & \checkmark & \checkmark & \checkmark & \checkmark & $\times$ & 0.0 & 0.0 & 20.1 & 37.5 & 14.7 & \underline{73.2} & 0.0 & 0.0 & 13.7 & \textbf{31.0} & \underline{34.3} & \textbf{60.2} \\
\bottomrule
\end{tabular}}
\end{table*}

\paragraph{Size of the set for cross-question rewards: a tradeoff between on-policy signal and robust risk estimation.}

The coverage and diversity rewards are computed against a set $\mathcal{S}$ of $(c_i, y_i)$ pairs (\Cref{sec:method-coverage}).
\Cref{tab:buffer-scope-ablation-id-ood} varies the size and scope of this set, ranging from a single GRPO group (32 completions) up to a 10{,}000-sample rolling buffer.
Increasing the size can yield more robust estimates of the abstention threshold, helping to avoid noisy rewards. 
However, a larger scope quickly becomes off-policy, as answers and confidence scores are generated by older versions of the model, which do not represent its current knowledge and confidence prediction capabilities.
For math, following \cite{damani2025beyond}, we use a GRPO group size of 32 (i.e.\ sample 32 answers to each question to estimate the baseline value), with 12 questions per GPU worker and 6 GPU workers in total.
Group-level pools (32) are too small to give a stable abstention threshold, and thus don't achieve coverage at low risk.
The per-worker and across-worker groups are the best (384 and 2{,}304 completions respectively). This represents the sweet spot, as they are on-policy yet provide sound abstention thresholds during training.
Per-worker is more efficient and achieves similar performance (higher ID but lower OOD) compared to across-workers, so we use that for the rest of our experiments.
One can accumulate a buffer of answer correctness labels and confidence scores from past updates.
However, larger sets perform poorly, likely due to off-policiness being too damaging and the added robustness small.

\begin{table*}[t]
\centering
\caption{
    \textbf{Analysis of the effect of the scope and size of the set of answers across which we compute rewards.} 
    In the math training regime, we study the tradeoff between smaller on-policy sets and larger sets that provide more robust abstention threshold estimates.
}
\label{tab:buffer-scope-ablation-id-ood}
\small
\setlength{\tabcolsep}{4pt}
\resizebox{\textwidth}{!}{\begin{tabular}{llrrrrr>{\color{gray}}rrrrrr>{\color{gray}}r}
\toprule
& & \multicolumn{6}{c}{\textbf{ID Math}} & \multicolumn{6}{c}{\textbf{OOD General Reasoning and Knowledge}} \\
\cmidrule(lr){3-8}\cmidrule(lr){9-14}
\textbf{Scope} & \textbf{Size} & \textbf{C@0} & \textbf{C@1} & \textbf{C@5} & \textbf{C@10} & \textbf{AURC$\downarrow$} & \textbf{Acc} & \textbf{C@0} & \textbf{C@1} & \textbf{C@5} & \textbf{C@10} & \textbf{AURC$\downarrow$} & \textbf{Acc} \\
\midrule
Per-group & 32 & 0.0 & 0.0 & 0.0 & 33.3 & 19.9 & 72.6 & 0.0 & 0.0 & 0.0 & 3.8 & 46.3 & 46.2 \\
Per-worker & 384 & \textbf{5.4} & \textbf{18.2} & \textbf{42.4} & \underline{52.9} & \textbf{12.5} & 73.1 & \textbf{2.9} & \textbf{5.4} & \underline{15.9} & \underline{26.4} & \underline{37.7} & 53.6 \\
Across-workers & 2{,}304 & \underline{2.2} & \underline{2.2} & \underline{28.7} & \textbf{54.3} & \underline{13.3} & 71.9 & \underline{1.5} & \underline{1.5} & \textbf{18.2} & \textbf{31.0} & \textbf{34.3} & 57.5 \\
Rolling buffer & 5{,}000 & 0.0 & 0.0 & 0.0 & 33.3 & 24.3 & \textbf{73.7} & 0.0 & 0.0 & 0.0 & 0.0 & 42.0 & \underline{57.7} \\
Rolling buffer & 10{,}000 & 0.0 & 0.0 & 0.0 & 33.3 & 33.9 & \underline{73.3} & 0.3 & 0.3 & 0.8 & 20.1 & 41.0 & \textbf{59.6} \\
\bottomrule
\end{tabular}}
\end{table*}

\section{Conclusion}
An LLM aware of the limits of its own knowledge and abilities can operate reliably in an autonomous manner.
We propose \ours, a reward added to RLVR that teaches the model to verbalize this confidence.
Unlike prior work that rewards each confidence against its own binary correctness label, \ours aggregates predicted confidences across a batch and reinforces each one relative to the others.
We demonstrate empirically across two training settings and various math, general knowledge, and reasoning benchmarks that \ours improves selective prediction, unlocking coverage at low risk without sacrificing overall accuracy.
Concretely, it enables a coverage at 5\% risk on math benchmarks of 42\% when existing verbalized baselines cannot surpass 1\%. 
We hope this motivates further work on learning to predict confidence for reliable reasoning, allowing control over risk in high stakes environments, as the coverage at low risk is still low in absolute terms compared to overall accuracy.

\clearpage
\renewcommand{\acksection}{\section*{Acknowledgments}}
\begin{ack}
This research was partially funded by an Alexander von Humboldt Professorship in
Multimodal Reliable AI, sponsored by the Federal Ministry of Research,
Technology, and Space (BMFTR), by a
LOEWE-Spitzen-Professur (LOEWE/4a//519/\allowbreak 05.00.002(0010)/93), and has
benefited from the Excellence Cluster ``Reasonable AI'' by the German Research
Foundation (Deutsche Forschungsgemeinschaft -- DFG) under Germany's Excellence
Strategy -- EXC-3057.
We gratefully acknowledge support from the hessian.AI Service Center (funded by
the Federal Ministry of Research, Technology and Space, BMFTR, grant no.
16IS22091) and the hessian.AI Innovation Lab (funded by the Hessian Ministry
for Digital Strategy and Innovation, grant no. S-DIW04/\allowbreak 0013/003).
\end{ack}

\bibliographystyle{plainnat}
\bibliography{references}

\clearpage
\appendix
\section{Supplementary Material}

\providecommand{\graycell}[1]{\textcolor{gray}{#1}}

\begin{table}[H]
\centering
\caption{Training datasets and evaluation benchmarks. Further details in \Cref{app:experimental-details}.}
\label{tab:datasets}
\footnotesize
\setlength{\tabcolsep}{4pt}
\resizebox{\linewidth}{!}{\begin{tabular}{llp{7.6cm}}
\toprule
\textbf{Dataset} & \textbf{Topic} & \textbf{Example} \\
\midrule
\multicolumn{3}{l}{\textit{Training Datasets}} \\
\midrule
Big-Math-Digits & Mixed-domain math & \textit{Q:} Given $X\!\sim\!B(n,0.8)$ and $D(X)\!=\!1.6$, find $n$. \textit{A:} 10 \\
HotpotQA & Wikipedia multi-hop QA & \textit{Q:} Were Scott Derrickson and Ed Wood of the same nationality? \textit{A:} yes \\
\midrule
\multicolumn{3}{l}{\textit{Test Benchmarks}} \\
\midrule
MATH-500 & Competition math & \textit{Q:} Convert $(0,3)$ from rectangular to polar coordinates. \textit{A:} \(\left(3,\frac{\pi}{2}\right)\) \\
GSM8K & Grade-school math & \textit{Q:} Janet's ducks lay 16 eggs; after eating 3 and baking with 4, how much money does she make selling the rest at \$2 each? \textit{A:} 18 \\
CommonsenseQA & Commonsense reasoning & \textit{Q:} A revolving door serves as security at a \underline{\hspace{0.5em}}? (A) bank \ldots \textit{A:} bank \\
GPQA & Graduate science & \textit{Q:} Two quantum states have lifetimes $10^{-9}$ s and $10^{-8}$ s; what resolvable energy gap fits? \textit{A:} $10^{-4}$ eV \\
SimpleQA & Short factual QA & \textit{Q:} Who received the IEEE Frank Rosenblatt Award in 2010? \textit{A:} Michio Sugeno \\
TriviaQA & Trivia QA & \textit{Q:} Which Lloyd Webber musical features the Phantom? \textit{A:} The Phantom of the Opera \\
\bottomrule
\end{tabular}
}
\end{table}

\subsection{Additional experimental details}
\label{app:experimental-details}

\paragraph{Training configuration.} We follow the training and evaluation protocol of \citet{damani2025beyond} (RLCR) unless stated otherwise.
We train Qwen2.5-7B with GRPO using the same answer-plus-confidence prompt format as RLCR.
The selected CARE runs use the per-worker reward scope: each GPU worker computes the cross-question rewards over its local batch of 12 prompts (16 for HotpotQA) with 32 completions each, for 384 completions (512 for HotpotQA) per reward pool.
Invalidly formatted completions are excluded from the confidence/correctness pool.
The coverage rewards use risk tolerances $\mathcal{T}=\{0,0.01,0.05,0.10,0.15,0.20,0.25\}$ and power $p=2$.
The entropy reward uses 100 equal-width bins on $[0,1]$, and the uniqueness reward uses tolerance $\delta=0.01$, so confidence values within one percentage point are treated as similar.
\cameraReadyEdit{Before weighting, we square the clipped entropy reward while preserving its sign, $u\mapsto u|u|$.}
We ablate using instead or in addition a 10-bin entropy reward and obtain worse results than with the 100-bin version.
These provide different diversity encouragements around bin boundaries even with similar granularity, which is relevant when the model tends to predict identical scores.
The selected reward weights are $0.5$ for format, $0.5$ for answer accuracy, $0.1$ for Brier, $0.1$ for uniqueness, $0.1$ for entropy, and $0.1$ total for the coverage rewards, split equally across risk levels.
For evaluation, tied confidence scores are accepted together, without using correctness to break ties.
We find that to avoid reward hacking, maintaining a balance between accuracy reward weight and other confidence related reward weights is quite relevant, increasing the coverage rewards carelessly can lead to scenarios where the model learns to give terrible answers with confidence 0, hence obtaining optimal confidence scores.

\paragraph{Answer extraction and judging.} Final answers are parsed from the standard RLCR answer-plus-confidence format.
Math benchmarks (Big-Math-Digits, MATH-500, GSM8K) are scored with the symbolic checker from \citet{damani2025beyond}.
The knowledge suite (CommonsenseQA, GPQA, SimpleQA, TriviaQA, HotpotQA) is scored with the LLM-as-judge pipeline of \citet{damani2025beyond}, using the same judge model and prompt.

\paragraph{Baselines.} For \textbf{RLVR}, \textbf{RLVR-Classifier}, and \textbf{RLCR}, we use the public checkpoints released by \citet{damani2025beyond}.
We implement the baselines for \textbf{CAR-WeightedBrier}, \textbf{CAR-MSCR}, and \textbf{BC-Beta} and follow the same training setting for these.
\cameraReadyEdit{For \textbf{RLSR} \citep{luo2026aligning}, we introduce their AURC-based reward into our setting, using full-parameter training and the same learning rate as all other methods.
For ORCE \citep{li2026orce}, we introduce their Spearman-correlation reward into our GRPO training, instead of training a separate confidence model with SFT and DPO as in the original work, as we seek to compare the reward designs in isolation.
We denote this adaptation as \textbf{ORCE-SC}.
}

\cameraReadyEdit{\paragraph{Gemma 4 E2B-it.}\label{app:gemma-details}
For Gemma 4 E2B-it, the training setup of \ours and RLCR is the same as for Qwen2.5-7B, except where stated below.
The vision and audio encoders are not used by our text-only tasks and are kept frozen.
Gemma results (\Cref{tab:rebuttal-gemma}) are averaged over two seeds.
Gemma generates considerably longer completions than Qwen2.5-7B (in our \ours runs, on average about 1.8k vs.\ 0.7k tokens during math training and 0.8k vs.\ 0.15k during HotpotQA training) and reaches the maximum completion length more often, particularly early in training (14.9\% vs.\ 1.1\% of completions over the first 50 HotpotQA steps).
Since longer sequences require more memory per training batch, Gemma runs out of GPU memory with the Qwen2.5-7B per-device training batch.
We therefore reduce the per-device training batch from 6 to 4 completions for math and from 8 to 4 for HotpotQA, and increase gradient accumulation from 64 to 96 and 128 steps, respectively.
This keeps the effective batch, the number of optimizer updates, and the set of completions over which the cross-question rewards are computed unchanged.
}

\subsection{Compute resources}
\label{app:compute-resources}

Training was run on NVIDIA H100 and A100 80GB GPUs.
Each Qwen2.5-7B run completed in under 24 hours; each Gemma 4 E2B-it run, with its smaller per-device batch, took 42 to 67 hours.

\subsection{Additional metrics}
\label{app:additional-metrics}
\label{app:additional-risk-level-results}
\label{app:rebuttal-calibration}

\paragraph{Additional coverage levels.}

\Cref{tab:math-id,tab:math-ood,tab:math-average} report coverage across the full set of evaluated risk levels for the math-trained models, and \Cref{tab:hotpot-id,tab:hotpot-ood} do the same for the HotpotQA-trained models.
Some entries remain zero even at high nominal risk levels because coverage is computed by thresholding the model's emitted confidence values, so a model with only a few tied confidence scores has only a few possible operating points.
For example, on Big-Math-Digits, RLVR emits only four confidence values.
The accepted sets obtained by thresholding these scores have error rates between 33.8\% and 35.7\%, so none satisfies the evaluated risk tolerances of up to 30\%.
Consequently, its coverage remains zero at all reported risk levels.

\begin{table}[t]
\centering
\caption{\textbf{Math, in-distribution} (Big-Math-Digits, 1{,}000 examples, \texttt{math-verify}).}
\label{tab:math-id}
\small
\setlength{\tabcolsep}{2.5pt}
\begin{tabular}{llrrrrrrrrr>{\color{gray}}r}
\toprule
\textbf{Model} & \textbf{Init} & \textbf{C@0} & \textbf{C@1} & \textbf{C@5} & \textbf{C@10} & \textbf{C@15} & \textbf{C@20} & \textbf{C@25} & \textbf{C@30} & \textbf{AURC\,$\downarrow$} & \textbf{Acc} \\
\midrule
\graycell{RLVR-Classifier \citep{damani2025beyond}} & \graycell{N/A} & \graycell{8.2} & \graycell{12.1} & \graycell{25.5} & \graycell{49.0} & \graycell{65.1} & \graycell{80.1} & \graycell{89.2} & \graycell{99.2} & \graycell{11.9} & \graycell{69.8} \\
\midrule
CAR-WeightedBrier \citep{xuan2026confidence} & Cold & 0.0 & 0.0 & 0.0 & 0.0 & 0.0 & 81.4 & 85.3 & \textbf{100.0} & 18.5 & 71.3 \\
CAR-MSCR \citep{xuan2026confidence} & Cold & \textbf{0.1} & \textbf{0.1} & 0.1 & 0.1 & 69.5 & 80.4 & 87.8 & 87.8 & 17.5 & 68.8 \\
BC-Beta \citep{wu2026behaviorally} & Cold & 0.0 & 0.0 & 0.0 & 51.9 & 71.6 & 81.5 & 86.6 & \textbf{100.0} & 13.9 & 70.0 \\
\midrule
RLSR \citep{luo2026aligning} & Cold & 0.0 & 0.0 & 0.0 & 0.0 & 0.0 & 83.7 & 85.6 & \textbf{100.0} & 20.0 & 70.7 \\
ORCE-SC \citep{li2026orce} & Cold & 0.0 & 0.0 & 0.0 & 0.0 & 0.0 & 0.0 & 0.0 & \textbf{100.0} & 26.2 & 71.6 \\
\midrule
RLVR \citep{damani2025beyond} & Cold & 0.0 & 0.0 & 0.0 & 0.0 & 0.0 & 0.0 & 0.0 & 0.0 & 34.7 & 66.2 \\
RLCR \citep{damani2025beyond} & SFT & 0.0 & 0.0 & 0.0 & 0.0 & 59.0 & 74.0 & 86.7 & \textbf{100.0} & 15.9 & 70.4 \\
RLCR \citep{damani2025beyond} & Cold & \textbf{0.1} & \textbf{0.1} & 0.1 & 0.1 & 0.1 & 68.3 & 81.4 & 97.6 & 20.1 & 68.7 \\
\textbf{\ours (Ours)} & Cold & \textbf{0.1} & \textbf{0.1} & \textbf{34.5} & \textbf{58.4} & \textbf{74.2} & \textbf{85.0} & \textbf{94.8} & \textbf{100.0} & \textbf{10.7} & \textbf{72.4} \\
\bottomrule
\end{tabular}
\end{table}

\begin{table}[t]
\centering
\caption{\textbf{Math-trained, out-of-distribution average} over CommonsenseQA, GPQA, SimpleQA, TriviaQA, and HotpotQA, scored with the LLM judge.}
\label{tab:math-ood}
\small
\setlength{\tabcolsep}{2.5pt}
\begin{tabular}{llrrrrrrrrr>{\color{gray}}r}
\toprule
\textbf{Model} & \textbf{Init} & \textbf{C@0} & \textbf{C@1} & \textbf{C@5} & \textbf{C@10} & \textbf{C@15} & \textbf{C@20} & \textbf{C@25} & \textbf{C@30} & \textbf{AURC\,$\downarrow$} & \textbf{Acc} \\
\midrule
\graycell{RLVR-Classifier \citep{damani2025beyond}} & \graycell{N/A} & \graycell{0.5} & \graycell{0.5} & \graycell{1.5} & \graycell{19.7} & \graycell{20.4} & \graycell{20.4} & \graycell{20.7} & \graycell{21.6} & \graycell{45.2} & \graycell{52.5} \\
\midrule
CAR-WeightedBrier \citep{xuan2026confidence} & Cold & 0.1 & 0.1 & 0.1 & 0.1 & 15.0 & 32.5 & 39.9 & 55.0 & 41.7 & 51.3 \\
CAR-MSCR \citep{xuan2026confidence} & Cold & 0.1 & 0.1 & 7.0 & 19.1 & \textbf{36.1} & 36.5 & 40.1 & 40.1 & 37.7 & 56.7 \\
BC-Beta \citep{wu2026behaviorally} & Cold & 0.0 & 0.0 & 11.6 & 19.6 & 32.3 & 36.1 & 39.8 & 40.3 & \textbf{37.6} & 55.7 \\
\midrule
RLSR \citep{luo2026aligning} & Cold & 0.0 & 0.0 & 0.0 & 0.0 & 0.0 & 0.0 & 13.0 & 13.0 & 51.0 & 43.9 \\
ORCE-SC \citep{li2026orce} & Cold & 0.0 & 0.0 & 0.0 & 20.0 & 20.0 & 20.0 & 38.1 & 40.0 & 40.9 & \textbf{56.8} \\
\midrule
RLVR \citep{damani2025beyond} & Cold & 0.0 & 0.0 & 0.0 & 0.0 & 20.0 & 20.0 & 20.0 & 20.0 & 50.1 & 49.5 \\
RLCR \citep{damani2025beyond} & SFT & 0.0 & 0.0 & 0.0 & 7.5 & 19.6 & 32.5 & 38.6 & 51.2 & 42.0 & 50.0 \\
RLCR \citep{damani2025beyond} & Cold & 0.0 & 0.0 & 4.9 & 20.0 & 32.2 & 38.3 & 40.1 & 53.3 & 38.7 & 56.6 \\
\textbf{\ours (Ours)} & Cold & \textbf{2.9} & \textbf{5.4} & \textbf{15.9} & \textbf{26.4} & 33.8 & \textbf{42.1} & \textbf{48.6} & \textbf{55.6} & 37.7 & 53.6 \\
\bottomrule
\end{tabular}
\end{table}

\begin{table}[t]
\centering
\caption{\textbf{Math average} over Big-Math-Digits, MATH-500, and GSM8K.}
\label{tab:math-average}
\small
\setlength{\tabcolsep}{4pt}
\resizebox{\linewidth}{!}{\begin{tabular}{llrrrrrrrrr>{\color{gray}}r}
\toprule
\textbf{Model} & \textbf{Init} & \textbf{C@0} & \textbf{C@1} & \textbf{C@5} & \textbf{C@10} & \textbf{C@15} & \textbf{C@20} & \textbf{C@25} & \textbf{C@30} & \textbf{AURC\,$\downarrow$} & \textbf{Acc} \\
\midrule
\graycell{RLVR-Classifier \citep{damani2025beyond}} & \graycell{N/A} & \graycell{4.6} & \graycell{5.9} & \graycell{33.6} & \graycell{52.7} & \graycell{59.2} & \graycell{75.4} & \graycell{85.5} & \graycell{92.7} & \graycell{12.8} & \graycell{73.7} \\
\midrule
CAR-WeightedBrier \citep{xuan2026confidence} & Cold & 0.0 & 0.0 & 0.0 & 33.3 & 33.3 & 60.5 & 61.8 & 66.7 & 20.6 & 71.8 \\
CAR-MSCR \citep{xuan2026confidence} & Cold & 0.0 & 0.0 & 0.0 & 33.4 & 56.5 & 60.1 & 62.6 & 62.6 & 19.2 & 72.0 \\
BC-Beta \citep{wu2026behaviorally} & Cold & 0.0 & 0.0 & 0.0 & 50.6 & 57.2 & 60.5 & 62.2 & 66.7 & 17.6 & 71.8 \\
\midrule
RLSR \citep{luo2026aligning} & Cold & 0.0 & 0.0 & 0.0 & 33.3 & 33.3 & 61.2 & 61.9 & 66.7 & 21.7 & 72.7 \\
ORCE-SC \citep{li2026orce} & Cold & 0.0 & 0.0 & 0.0 & 33.3 & 33.3 & 33.3 & 33.3 & 66.7 & 23.6 & \textbf{75.0} \\
\midrule
RLVR \citep{damani2025beyond} & Cold & 0.0 & 0.0 & 0.0 & 33.3 & 33.3 & 33.3 & 33.3 & 33.3 & 31.0 & 70.3 \\
RLCR \citep{damani2025beyond} & SFT & 0.0 & 0.0 & 0.0 & 33.3 & 53.0 & 58.0 & 82.2 & 89.9 & 16.3 & 72.0 \\
RLCR \citep{damani2025beyond} & Cold & 0.0 & 0.0 & 0.8 & 32.9 & 34.3 & 57.0 & 61.4 & 91.2 & 19.3 & 73.1 \\
\textbf{\ours (Ours)} & Cold & \textbf{5.4} & \textbf{18.2} & \textbf{42.4} & \textbf{52.9} & \textbf{66.8} & \textbf{76.4} & \textbf{84.9} & \textbf{91.4} & \textbf{12.5} & 73.1 \\
\bottomrule
\end{tabular}}
\end{table}

\begin{table}[t]
\centering
\caption{\cameraReadyEdit{HotpotQA-trained models, in-distribution (HotpotQA distractor, 1{,}000 examples, Llama judge).}}
\label{tab:hotpot-id}
\small
\setlength{\tabcolsep}{3.5pt}
\begin{tabular}{lrrrrrrrrr>{\color{gray}}r}
\toprule
\textbf{Model} & \textbf{C@0} & \textbf{C@1} & \textbf{C@5} & \textbf{C@10} & \textbf{C@15} & \textbf{C@20} & \textbf{C@25} & \textbf{C@30} & \textbf{AURC\,$\downarrow$} & \textbf{Acc} \\
\midrule
\graycell{RLVR-Classifier \citep{damani2025beyond}} & \graycell{1.6} & \graycell{1.6} & \graycell{14.2} & \graycell{92.8} & \graycell{100.0} & \graycell{100.0} & \graycell{100.0} & \graycell{100.0} & \graycell{6.5} & \graycell{87.7} \\
\midrule
CAR-WeightedBrier \citep{xuan2026confidence} & 0.0 & 0.0 & 0.0 & 90.0 & \textbf{100.0} & \textbf{100.0} & \textbf{100.0} & \textbf{100.0} & 6.4 & 85.8 \\
CAR-MSCR \citep{xuan2026confidence} & 0.0 & 0.0 & 0.0 & 89.6 & \textbf{100.0} & \textbf{100.0} & \textbf{100.0} & \textbf{100.0} & 7.2 & 87.0 \\
BC-Beta \citep{wu2026behaviorally} & 0.1 & 0.1 & 0.1 & 89.7 & 89.7 & \textbf{100.0} & \textbf{100.0} & \textbf{100.0} & 7.2 & 81.9 \\
\midrule
RLSR \citep{luo2026aligning} & 0.0 & 0.0 & 0.0 & 85.6 & \textbf{100.0} & \textbf{100.0} & \textbf{100.0} & \textbf{100.0} & 8.6 & 86.6 \\
ORCE-SC \citep{li2026orce} & 0.0 & 0.0 & 0.0 & 0.0 & \textbf{100.0} & \textbf{100.0} & \textbf{100.0} & \textbf{100.0} & 10.7 & 85.6 \\
\midrule
RLVR \citep{damani2025beyond} & 0.0 & 0.0 & 0.0 & 0.0 & \textbf{100.0} & \textbf{100.0} & \textbf{100.0} & \textbf{100.0} & 12.3 & \textbf{87.7} \\
RLCR \citep{damani2025beyond} & 2.4 & 2.4 & 34.7 & 87.5 & \textbf{100.0} & \textbf{100.0} & \textbf{100.0} & \textbf{100.0} & 6.1 & 86.1 \\
\textbf{\ours (Ours)} & \textbf{6.1} & \textbf{6.1} & \textbf{74.7} & \textbf{93.5} & \textbf{100.0} & \textbf{100.0} & \textbf{100.0} & \textbf{100.0} & \textbf{4.4} & 87.6 \\
\bottomrule
\end{tabular}
\end{table}

\begin{table}[t]
\centering
\caption{\textbf{HotpotQA-trained, out-of-distribution average} over MATH-500, GSM8K, CommonsenseQA, GPQA, SimpleQA, and TriviaQA.}
\label{tab:hotpot-ood}
\small
\setlength{\tabcolsep}{3.5pt}
\begin{tabular}{lrrrrrrrrr>{\color{gray}}r}
\toprule
\textbf{Model} & \textbf{C@0} & \textbf{C@1} & \textbf{C@5} & \textbf{C@10} & \textbf{C@15} & \textbf{C@20} & \textbf{C@25} & \textbf{C@30} & \textbf{AURC\,$\downarrow$} & \textbf{Acc} \\
\midrule
\graycell{RLVR-Classifier \citep{damani2025beyond}} & \graycell{0.4} & \graycell{0.4} & \graycell{6.0} & \graycell{17.5} & \graycell{17.9} & \graycell{33.7} & \graycell{34.7} & \graycell{36.9} & \graycell{42.0} & \graycell{53.9} \\
\midrule
CAR-WeightedBrier \citep{xuan2026confidence} & 0.0 & 0.0 & 16.5 & 32.4 & 33.4 & \textbf{44.0} & 46.0 & 46.0 & 35.6 & 57.3 \\
CAR-MSCR \citep{xuan2026confidence} & 0.0 & \textbf{7.4} & 14.8 & 31.4 & 32.2 & 42.7 & 44.3 & 44.3 & 36.2 & 56.0 \\
BC-Beta \citep{wu2026behaviorally} & 0.9 & 5.6 & 14.8 & \textbf{34.0} & 40.7 & 42.5 & \textbf{50.0} & 50.0 & \textbf{30.5} & \textbf{63.4} \\
\midrule
RLSR \citep{luo2026aligning} & 0.0 & 0.0 & 14.5 & 32.8 & 33.3 & 33.3 & 33.3 & 45.6 & 37.7 & 58.8 \\
ORCE-SC \citep{li2026orce} & 0.0 & 0.0 & 14.9 & 16.7 & 31.7 & 32.5 & 43.6 & 45.4 & 38.5 & 55.2 \\
\midrule
RLVR \citep{damani2025beyond} & 0.0 & 0.0 & 0.0 & 16.7 & 16.7 & 33.3 & 33.3 & 33.3 & 46.0 & 53.9 \\
RLCR \citep{damani2025beyond} & 0.1 & 0.1 & 12.3 & 26.9 & 33.7 & 38.3 & 40.1 & 44.5 & 36.6 & 56.2 \\
\textbf{\ours (Ours)} & \textbf{3.1} & 6.4 & \textbf{23.6} & 33.1 & \textbf{40.8} & 42.2 & 44.6 & \textbf{50.7} & 33.2 & 58.0 \\
\bottomrule
\end{tabular}
\end{table}

\cameraReadyEdit{\paragraph{Additional ranking and calibration metrics.}
We complement coverage at different risk levels and AURC with AUROC, expected calibration error (ECE), and Brier score.

\textbf{Confidence ranking.}
AUROC measures whether correct answers receive higher confidence than incorrect answers, considering all correct--incorrect pairs~\citep{fawcett2006introduction}.
Let $I_1$ and $I_0$ denote the sets of correct and incorrect completions:
\begin{equation}
\mathrm{AUROC}=\frac{1}{|I_1||I_0|}
\sum_{i\in I_1}\sum_{j\in I_0}
\left(\indicator[c_i>c_j]+\tfrac12\indicator[c_i=c_j]\right).
\end{equation}
Each pair contributes equally.
In contrast, AURC evaluates the error rates of accepted sets obtained by varying a confidence threshold, giving greater importance to errors among the highest-confidence answers.
For example, consider four completions with confidence scores $(0.9,0.8,0.7,0.6)$.
If their correctness labels in this order are (correct, wrong, wrong, correct), accuracy and AUROC are both 50\%, and AURC is 41.7\%.
With labels (wrong, correct, correct, wrong), accuracy and AUROC remain 50\%, but AURC increases to 58.3\%.
Both examples correctly rank two of the four correct--incorrect pairs.
However, the second places an incorrect answer first, increasing the risk of accepting only the highest-confidence answers and consequently worsening AURC.
\Cref{tab:rebuttal-calibration} shows that \ours achieves the highest AUROC in all training and evaluation settings, complementing the coverage results with evidence of better ranking of correct versus incorrect answers.

\textbf{Calibration.}
Calibration alone does not ensure useful selective prediction.
A model that assigns every answer a confidence equal to its overall accuracy has zero empirical ECE, but cannot distinguish which answers to accept.
Even a model that always answers incorrectly and always predicts zero confidence is perfectly calibrated, despite providing no useful answers.
Nevertheless, alongside good coverage, calibration can be helpful when abstention thresholds cannot be tuned on labeled validation data and confidence values must be interpreted directly.

Because \ours trains confidence for coverage rather than probability calibration, its confidence scale need not match probabilities of correctness.
A simple post-hoc calibration function can adjust this scale.
Following Platt-style scaling~\citep{platt1999probabilistic,guo2017calibration}, we fit
\begin{equation}
c'_i=\sigma\!\left(a\,\operatorname{logit}(c_i)+b\right),\qquad a>0.
\end{equation}
The positive slope preserves confidence ordering, allowing probability calibration without changing coverage at any risk level or AURC.
We fit the two parameters by minimizing binary negative log-likelihood on 1,000 Math training examples or 500 HotpotQA training examples.
Before applying the logit, we clip confidence scores to $[10^{-6},1-10^{-6}]$.
This is necessary because the logit is infinite at $c=0$ and $c=1$.
Clipping moves these endpoint scores slightly inside the interval so that the calibration function can be fitted and evaluated.
The fitted function is applied unchanged to ID and OOD evaluation examples.

We evaluate the resulting probabilities using ECE~\citep{guo2017calibration} and the binary Brier score~\citep{brier1950verification}:
\begin{align}
\mathrm{ECE}&=\sum_{m=1}^{20}\frac{|B_m|}{N}
\left|\frac{1}{|B_m|}\sum_{i\in B_m}y_i
-\frac{1}{|B_m|}\sum_{i\in B_m}c'_i\right|,\\
\mathrm{Brier}&=\frac1N\sum_{i=1}^{N}(c'_i-y_i)^2,
\end{align}
where $B_m$ contains examples whose recalibrated confidence falls in the $m$-th equal-width bin, with empty bins omitted.
ECE measures agreement between average confidence and accuracy within bins, while Brier measures squared probability error.
Lower values are better for both.
\Cref{tab:rebuttal-calibration} shows that \ours achieves the lowest or tied-lowest Brier score after recalibration in all training and evaluation settings.
RLCR achieves lower ECE on ID evaluation, but this advantage does not transfer to OOD, where \ours improves or matches its ECE.
}

\begin{table}[t]
\centering\small
\caption{\cameraReadyEdit{\textbf{\ours provides the best overall confidence ranking and calibration.} Only in the ID scenario does RLCR achieve a better ECE than \ours, which does not generalize to OOD. \textcolor{gray}{RLVR-Classifier} is the only method not using verbalized confidence prediction from an LLM. \textbf{Bold} and \underline{underlining} mark the best and second-best results among methods with verbalized confidence within each training setting, including ties.}}
\label{tab:rebuttal-calibration}
\resizebox{\linewidth}{!}{\begin{tabular}{llrrrrrr}
\toprule
 & & \multicolumn{3}{c}{\textbf{ID}} & \multicolumn{3}{c}{\textbf{OOD}} \\
\cmidrule(lr){3-5}\cmidrule(lr){6-8}
\textbf{Training} & \textbf{Method} & \textbf{AUROC\,$\uparrow$} & \textbf{Brier\,$\downarrow$} & \textbf{ECE\,$\downarrow$} & \textbf{AUROC\,$\uparrow$} & \textbf{Brier\,$\downarrow$} & \textbf{ECE\,$\downarrow$} \\
\midrule
 & \textcolor{gray}{RLVR-Classifier \citep{damani2025beyond}} & \textcolor{gray}{78.2} & \textcolor{gray}{.149} & \textcolor{gray}{9.3} & \textcolor{gray}{54.6} & \textcolor{gray}{.343} & \textcolor{gray}{34.3} \\
\cmidrule(lr){2-8}
 & RLVR \citep{damani2025beyond} & 46.5 & .209 & \underline{12.3} & 51.6 & .283 & \underline{25.7} \\
\raisebox{0.5\normalbaselineskip}[0pt][0pt]{Math} & RLCR \citep{damani2025beyond} & \underline{67.8} & \underline{.166} & \textbf{8.7} & \underline{61.5} & \underline{.278} & \underline{25.7} \\
 & \textbf{\ours} & \textbf{83.7} & \textbf{.148} & 12.5 & \textbf{67.3} & \textbf{.233} & \textbf{22.1} \\
\midrule
 & \textcolor{gray}{RLVR-Classifier \citep{damani2025beyond}} & \textcolor{gray}{72.1} & \textcolor{gray}{.102} & \textcolor{gray}{5.8} & \textcolor{gray}{58.3} & \textcolor{gray}{.320} & \textcolor{gray}{30.1} \\
\cmidrule(lr){2-8}
 & RLVR \citep{damani2025beyond} & 50.0 & \underline{.125} & 13.1 & 50.1 & .291 & \underline{27.9} \\
\raisebox{0.5\normalbaselineskip}[0pt][0pt]{HotpotQA} & RLCR \citep{damani2025beyond} & \underline{78.8} & \textbf{.098} & \textbf{1.9} & \underline{67.7} & \underline{.273} & \textbf{27.4} \\
 & \textbf{\ours} & \textbf{80.7} & \textbf{.098} & \underline{8.0} & \textbf{71.4} & \textbf{.269} & \textbf{27.4} \\
\bottomrule
\end{tabular}}
\end{table}

\subsection{Analysis of confidence distributions of other baselines}
\label{app:additional-analyses}

\begin{figure*}[t]
\centering
\includegraphics[width=\textwidth]{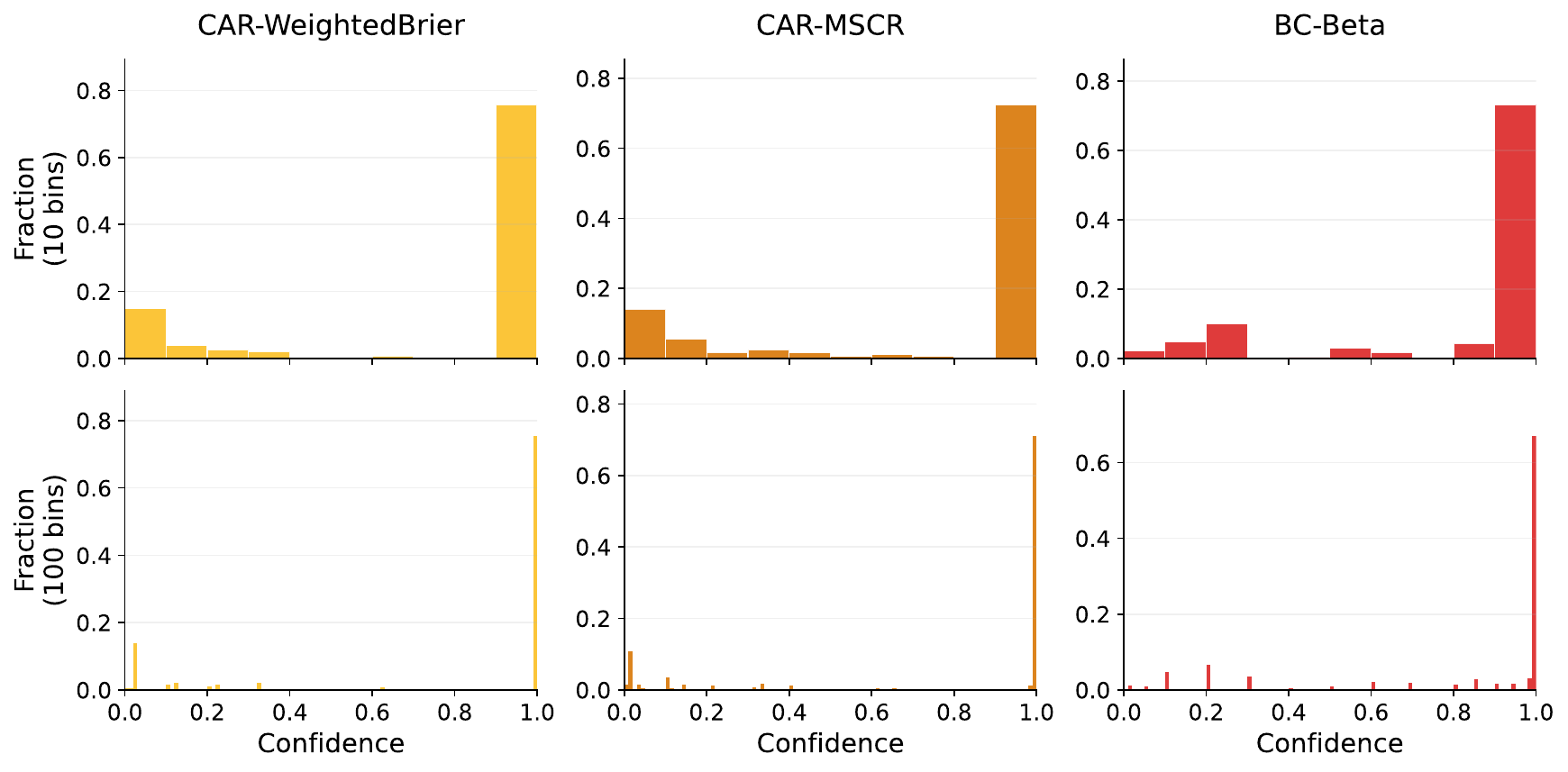}
\caption{\textbf{Big-Math-Digits confidence distributions for the remaining baselines} (CAR-WeightedBrier, CAR-MSCR, BC-Beta), extending \Cref{fig:math-confidence-distributions}. Mass concentrates on a small number of bins, in contrast to \ours's finer-grained range.}
\label{fig:appendix-baseline-confidence-distributions}
\end{figure*}

\Cref{fig:appendix-baseline-confidence-distributions} extends the Big-Math-Digits confidence-distribution analysis of \Cref{fig:math-confidence-distributions} to the remaining pointwise confidence-learning baselines, CAR-WeightedBrier, CAR-MSCR, and BC-Beta.
All three follow the same overall trend as RLCR: mass is heavily concentrated in the high-confidence regime, so these methods remain overconfident in aggregate.
They differ from RLCR in two minor ways that the figure makes visible.
First, the CAR variants place a small but non-trivial amount of mass near zero, and BC-Beta places a smaller bump there as well, whereas RLCR almost never predicts very low confidence.
Second, beyond the single-decimal values that dominate the verbalized baselines, these methods occasionally use additional values such as $0.95$ or $0.85$, slightly enlarging the effective vocabulary of confidence scores.
The set of distinct values used remains very small compared to \ours, however, so the fine-grained histogram is still dominated by a handful of round-number bins rather than the broader continuous range produced by our method.

\subsection{Additional ablations}
\label{app:supp-ablations}

\Cref{tab:reward-geometry-ablation-master} compares exponents applied to the confidence distance from the abstention threshold. Squaring this distance improves ID coverage and AURC, while the linear reward performs better on OOD evaluation.

\begin{table}[H]
\centering
\caption{\textbf{Reward-geometry ablation:} the exponent controls how strongly larger confidence distances are weighted relative to smaller ones.
ID averages Big-Math-Digits, MATH-500, and GSM8K. OOD averages CommonsenseQA, GPQA, SimpleQA, TriviaQA, and HotpotQA.
}
\label{tab:reward-geometry-ablation-master}
\small
\setlength{\tabcolsep}{4pt}
\resizebox{\textwidth}{!}{\begin{tabular}{lcccc>{\color{gray}}ccccc>{\color{gray}}c}
\toprule
& \multicolumn{5}{c}{\textbf{ID}} & \multicolumn{5}{c}{\textbf{OOD}} \\
\cmidrule(lr){2-6}\cmidrule(lr){7-11}
\textbf{Reward transform} & \textbf{C@0} & \textbf{C@1} & \textbf{C@5} & \textbf{AURC\,$\downarrow$} & \textbf{Acc} & \textbf{C@0} & \textbf{C@1} & \textbf{C@5} & \textbf{AURC\,$\downarrow$} & \textbf{Acc} \\
\midrule
Linear ($p{=}1.0$) & 0.6 & 0.6 & \underline{36.1} & 13.7 & \underline{72.7} & \underline{1.2} & \underline{4.3} & \textbf{17.1} & \textbf{34.9} & \textbf{58.8} \\
Decay exponent ($p{=}0.8$) & \underline{3.4} & \underline{9.7} & 35.3 & \underline{13.5} & 72.3 & 0.4 & 0.4 & \underline{16.3} & \underline{35.8} & \underline{56.3} \\
Squared ($p{=}2.0$) & \textbf{5.4} & \textbf{18.2} & \textbf{42.4} & \textbf{12.5} & \textbf{73.1} & \textbf{2.9} & \textbf{5.4} & 15.9 & 37.7 & 53.6 \\
\bottomrule
\end{tabular}}
\end{table}

\cameraReadyEdit{\subsection{Brier reward and training progression}
\label{app:rebuttal-brier}

During the first 50 Math training steps, the average absolute advantage contribution before GRPO reward scaling is $0.023$ from the Brier reward and $0.011$ from the combined coverage rewards, more than twice as much.
The ratio falls to $1.08$ over steps 51--100, $1.01$ over 101--150, and $0.96$ over 151--208.

The larger initial contribution does not necessarily mean that \ours's coverage rewards would fail without the Brier term.
In fact, on Math, the no-Brier ablation improves ID AURC from $12.5\%$ to $12.1\%$ and OOD AURC from $37.7\%$ to $36.9\%$, as shown in \Cref{tab:reward-design-ablation-id-ood}.
ID accuracy also improves from $73.1\%$ to $73.9\%$, and OOD accuracy from $53.6\%$ to $56.5\%$.
}

\clearpage
\newpage
\section*{NeurIPS Paper Checklist}

\begin{enumerate}

\item {\bf Claims}
    \item[] Question: Do the main claims made in the abstract and introduction accurately reflect the paper's contributions and scope?
    \item[] Answer: \answerYes{}
    \item[] Justification: The abstract and introduction state the paper's core contribution, the cross-question reward mechanism, and the main empirical improvements. The results and conclusion also state that, although large improvements are observed with the proposed method, gains are not universal and difficult low-risk and OOD settings remain unsolved.

\item {\bf Limitations}
    \item[] Question: Does the paper discuss the limitations of the work performed by the authors?
    \item[] Answer: \answerYes{}
    \item[] Justification: The results discuss rare cases where alternative methods that focus more directly on preserving or improving accuracy can outperform \ours, especially on some OOD settings.

\item {\bf Theory assumptions and proofs}
    \item[] Question: For each theoretical result, does the paper provide the full set of assumptions and a complete (and correct) proof?
    \item[] Answer: \answerNA{}
    \item[] Justification: The paper proposes and evaluates a training reward and does not claim formal theoretical results requiring proofs.

\item {\bf Experimental result reproducibility}
    \item[] Question: Does the paper fully disclose all the information needed to reproduce the main experimental results of the paper to the extent that it affects the main claims and/or conclusions of the paper (regardless of whether the code and data are provided or not)?
    \item[] Answer: \answerYes{}
    \item[] Justification: The paper specifies the datasets, evaluation benchmarks, scoring pipelines, reward design, selected hyperparameters, effective batch geometry, public baseline sources, and the additional ablations needed to reproduce the main claims.

\item {\bf Open access to data and code}
    \item[] Question: Does the paper provide open access to the data and code, with sufficient instructions to faithfully reproduce the main experimental results, as described in supplemental material?
    \item[] Answer: \answerNo{}
    \item[] Justification: The datasets and baseline code used in the work are public, and our implementation is based heavily on the public RLCR codebase. Necessary details to implement our method within the public code repository are presented in the paper, but our version of the code will only be made public at a later date.

\item {\bf Experimental setting/details}
    \item[] Question: Does the paper specify all the training and test details (e.g., data splits, hyperparameters, how they were chosen, type of optimizer, etc.) necessary to understand the results?
    \item[] Answer: \answerYes{}
    \item[] Justification: The main text and appendix specify the model family, datasets, prompt format, optimizer family, learning rates, schedules, warmup, batch sizes, context lengths, generation counts, selected reward weights, evaluation procedure, and the ablations used to validate the design.

\item {\bf Experiment statistical significance}
    \item[] Question: Does the paper report error bars suitably and correctly defined or other appropriate information about the statistical significance of the experiments?
    \item[] Answer: \answerYes{}
    \item[] Justification: The paper evaluates consistently across two distinct training scenarios and multiple in-distribution and out-of-distribution benchmarks, and all reported evaluations use deterministic temperature-0 decoding to reduce sampling noise. Together, this gives a robust picture of the empirical effects even without separate confidence intervals or significance tests.

\item {\bf Experiments compute resources}
    \item[] Question: For each experiment, does the paper provide sufficient information on the computer resources (type of compute workers, memory, time of execution) needed to reproduce the experiments?
    \item[] Answer: \answerYes{}
    \item[] Justification: The appendix states that training was run on NVIDIA H100 and A100 80GB GPUs and that the reported training runs completed in under 24 hours for Qwen2.5-7B, and in up to 67 hours for the Gemma models.

\item {\bf Code of ethics}
    \item[] Question: Does the research conducted in the paper conform, in every respect, with the NeurIPS Code of Ethics \url{https://neurips.cc/public/EthicsGuidelines}?
    \item[] Answer: \answerYes{}
    \item[] Justification: The work studies confidence estimation and selective prediction using existing benchmark data and models. The draft does not describe data collection from human subjects or other procedures that would require a deviation from the NeurIPS Code of Ethics.

\item {\bf Broader impacts}
    \item[] Question: Does the paper discuss both potential positive societal impacts and negative societal impacts of the work performed?
    \item[] Answer: \answerYes{}
    \item[] Justification: The introduction motivates positive uses in assistive and automation settings, discussing both the benefits of better abstention and the risks of over-trust and unsafe over-automation under distribution shift.

\item {\bf Safeguards}
    \item[] Question: Does the paper describe safeguards that have been put in place for responsible release of data or models that have a high risk for misuse (e.g., pretrained language models, image generators, or scraped datasets)?
    \item[] Answer: \answerNA{}
    \item[] Justification: The paper does not release a new high-risk pretrained model, image generator, or scraped dataset.

\item {\bf Licenses for existing assets}
    \item[] Question: Are the creators or original owners of assets (e.g., code, data, models), used in the paper, properly credited and are the license and terms of use explicitly mentioned and properly respected?
    \item[] Answer: \answerNo{}
    \item[] Justification: The draft credits the relevant prior work, datasets, and model families, but it does not yet explicitly enumerate the licenses or terms of use for each external dataset, model, and code asset used.

\item {\bf New assets}
    \item[] Question: Are new assets introduced in the paper well documented and is the documentation provided alongside the assets?
    \item[] Answer: \answerNA{}
    \item[] Justification: The paper does not introduce a new dataset, benchmark, model checkpoint, or code package as a new asset.

\item {\bf Crowdsourcing and research with human subjects}
    \item[] Question: For crowdsourcing experiments and research with human subjects, does the paper include the full text of instructions given to participants and screenshots, if applicable, as well as details about compensation (if any)?
    \item[] Answer: \answerNA{}
    \item[] Justification: The experiments use existing benchmarks and automated or LLM-based evaluation pipelines; the draft does not report new crowdsourcing or human-subjects studies.

\item {\bf Institutional review board (IRB) approvals or equivalent for research with human subjects}
    \item[] Question: Does the paper describe potential risks incurred by study participants, whether such risks were disclosed to the subjects, and whether Institutional Review Board (IRB) approvals (or an equivalent approval/review based on the requirements of your country or institution) were obtained?
    \item[] Answer: \answerNA{}
    \item[] Justification: The draft does not involve new human-subjects research, so IRB or equivalent review is not applicable.

\item {\bf Declaration of LLM usage}
    \item[] Question: Does the paper describe the usage of LLMs if it is an important, original, or non-standard component of the core methods in this research? Note that if the LLM is used only for writing, editing, or formatting purposes and does \emph{not} impact the core methodology, scientific rigor, or originality of the research, declaration is not required.
    \item[] Answer: \answerNA{}
    \item[] Justification: the LLM is used only for writing, editing, or formatting purposes and does \emph{not} impact the core methodology, scientific rigor, or originality of the research.

\end{enumerate}

\end{document}